# A Quantum Tunneling and Bio-Phototactic Driven Enhanced Dwarf Mongoose Optimizer for UAV Trajectory Planning and Engineering Problem

Mingyang Yu[a], Haorui Yang[a], Kangning An[a], Xinjian Wei[a], Xiaoxuan Xu[a], Jing Xu[a],*

[a]College of Artificial Intelligence, Nankai University, Tianjin, 300350, China
*Corresponding author

Mingyang Yu
Ph.D., Student
College of Artificial Intelligence,
Nankai University, Tianjin, China 300350
Email: 1120240312@mail.nankai.edu.cn

Haorui Yang
Postgraduate Student
College of Artificial Intelligence,
Nankai University, Tianjin, China 300350
Email: 2120240645@mail.nankai.edu.cn

Kangning An
Postgraduate Student
College of Artificial Intelligence,
Nankai University, Tianjin, China 300350
Email: 2120240547@mail.nankai.edu.cn

Xinjian Wei
Ph.D., Student
College of Artificial Intelligence,
Nankai University, Tianjin, China 300350
Email: wxj_wace@mail.nankai.edu.cn

Xiaoxuan Xu
Ph.D.
College of Artificial Intelligence,
Nankai University, Tianjin, China 300350
Email: xuxx@nankai.edu.cn

Jing Xu
Ph.D., Professor
College of Artificial Intelligence,
Nankai University, Tianjin, China 300350
Email: xujing@nankai.edu.cn

**Abstract:** Path planning in high-dimensional spaces faces NP-hard challenges due to rugged fitness landscapes and exponentially growing search space. While metaheuristic (MH) algorithms like Dwarf Mongoose Optimization (DMO) offer solutions, their slow convergence and limited global exploration hinder practical applications. To address these challenges, an Enhanced Multi-Strategy Dwarf Mongoose Optimization Algorithm (EDMO) is proposed. EDMO integrates two novel strategies: a Dynamic Quantum Tunneling Optimization Strategy (DQTOS), which enables particles to probabilistically traverse local optima barriers and enhance global exploration, and a Bio-phototactic Dynamic Focusing Search Strategy (BDFSS), which draws inspiration from microbial phototactic behavior to dynamically refine local search trajectories. Additionally, EDMO incorporates an Orthogonal Lens Opposition-Based Learning (OLOBL) strategy combined with orthogonal experimental design to further improve exploration efficiency by leveraging dominant dimensional information. To demonstrate the superior optimization precision, speed, and robustness of EDMO, we conduct tests on 39 benchmark functions, including those from CEC2017 and CEC2020. EDMO is compared against three categories of well-established algorithms: (1) competition winners such as CMA-ES, LSHADE_cnEpsin, and LSHADE; (2) advanced algorithms including QANA, SMO, IAO, IPO, MSMA, GWCA, OMA, and GSA; and (3) the original DMO and its variants, CO-DWO and BDMSAO. Experimental results demonstrate that EDMO consistently outperforms these algorithms in terms of convergence speed, solution stability, and optimization accuracy, with statistical significance confirmed by the Wilcoxon signed-rank test and Friedman mean rank test. Furthermore, real-world validations on engineering design problems and UAV three-dimensional path planning tasks confirm the superior robustness, adaptability, and practical applicability of EDMO in complex optimization environments.

## 1 Introduction

Unmanned Aerial Vehicles (UAVs) are widely utilized across various fields due to their high degree of autonomy, cost-effectiveness, rapid response capabilities, flexible deployment, and significant scalability [1, 2]. Over the past few decades, technological advancements have greatly enhanced the practicality of UAVs for a diverse range of applications, including agricultural monitoring [3, 4], aerial photography [5], search and rescue operations [6], and logistical support [7]. However, as these applications have expanded, the limitations of traditional two-dimensional flight paths have become increasingly evident, underscoring the need for sophisticated three-dimensional path planning to navigate diverse terrains and obstacles effectively [8]. This complexity has classified the task as a classic NP-hard problem, highlighting its computational demands [9, 10].

In the field of UAV three-dimensional path planning, various advanced techniques and algorithms are available. While some systems still utilize traditional optimization methods, the limitations of these classic deterministic algorithms become increasingly apparent as the scale and complexity of problems grow [11]. Particularly for NP-hard problems, these algorithms often struggle to find the best or near-optimal solutions within a limited timeframe [12]. Given the diverse environments, ever-changing obstacles, and numerous constraints that UAV three-dimensional path planning must navigate, the solution search space becomes cumbersome, with the number of potential solutions potentially surging dramatically [1]. Thus, relying on traditional deterministic algorithms to evaluate all possible flight paths within polynomial time becomes somewhat unrealistic [13].

To address these challenges, metaheuristic (MH) algorithms, inspired by natural biological behaviors, play a crucial role in overcoming the difficulties associated with UAV path planning. Renowned for their user-friendliness and adaptability, these algorithms have demonstrated significant effectiveness in fields such as image segmentation and path planning [14, 15]. This broad applicability highlights their value in tackling diverse optimization challenges, particularly in scenarios where traditional deterministic methods fall short.

However, as the complexity of UAV 3D path planning tasks increases, traditional optimization methods,

including some MH algorithms, face limitations—especially when rapid identification of optimal solutions is required in NP-hard scenarios [16-18]. The No Free Lunch (NFL) Theorem emphasizes that there is no universally perfect algorithm for every problem [19]. This underscores the importance of customizing algorithms for specific tasks. Therefore, developing MH algorithms tailored for 3D UAV path planning is crucial for improving efficiency, in line with the NFL Theorem, and highlights the need for specialized solutions to address complex problems.

In this context, the Dwarf Mongoose Optimization Algorithm (DMO) is a swarm intelligence optimization method inspired by the foraging behavior of mongooses searching for food and exploring mounds for resting. Introduced by Jeffrey O. Agushaka and colleagues in 2022 [20], DMO is noted for its simple structure, fewer parameters, and strong prospecting and exploration capabilities, making it suitable for solving optimization problems in areas such as high-dimensional feature selection and data clustering. However, despite these advantages, the DMO algorithm has some drawbacks, including slower convergence speed and a tendency to get trapped in local optima [21]. Additionally, focusing solely on flight distance and time may oversimplify the UAV path planning problem, hindering the ability to achieve optimal real-world flight paths. To date, only a few researchers have explored applying DMO to UAV three-dimensional path planning. For all optimization algorithms, achieving a proper balance between exploration and exploitation is crucial for realizing ideal flight paths [22, 23]. In summary, DMO is a relatively new algorithm that requires further research and refinement to effectively address the real-world challenges of three-dimensional path planning.

To address the limitations of the DMO algorithm, this research introduces an enhanced variant, EDMO, which incorporates a series of strategic improvements to boost performance. First, the Orthogonal Lens Opposition-Based Learning (OLOBL) strategy is employed to generate reflected opposition solutions, allowing unexplored regions of the search space to be covered, thereby enhancing global search capability and reducing the risk of local optima. Second, to overcome the inherent tendency of conventional swarm-based algorithms to become trapped in local optima in high-dimensional landscapes, a Dynamic Quantum Tunneling Optimization Strategy (DQTOS) is introduced. Inspired by quantum tunneling phenomena, this strategy allows particles to probabilistically traverse energy barriers, thereby enhancing the algorithm's capability to escape local optima and strengthen global search performance. Third, to further improve local exploitation accuracy and prevent stagnation, a Bio-phototactic Dynamic Focusing Search Strategy (BDFSS) is proposed. Drawing inspiration from microbial phototactic behavior, this mechanism establishes a dynamic mapping between fitness landscapes and virtual light intensity, enabling adaptive focusing on promising regions. Empirical evaluations on 39 benchmark functions from the CEC2017 and CEC2020 suites demonstrate that EDMO significantly outperforms state-of-the-art algorithms in terms of convergence speed, global search capacity, and solution robustness. In addition, preliminary validations in real-world UAV 3D trajectory planning tasks confirm its effectiveness in handling complex, high-dimensional engineering scenarios. The principal contributions are summarized as follows:

1) The OLOBL strategy is adopted, introducing an innovative opposition-based learning mechanism. The orthogonal table is designed to map the current solution to unexplored regions of the search space, generating inverse solutions. This enhances the algorithm's ability to escape local optima and ensures a more focused and efficient global search.
2) A DQTOS is proposed, inspired by quantum tunneling phenomena. This strategy endows particles with quantum tunneling capabilities to penetrate the energy barriers of local optima, thereby enhancing global search performance in complex optimization scenarios.
3) Inspired by the self-organizing properties, adaptive mechanisms, and efficient foraging characteristics inherent in microbial phototactic behavior observed in natural systems, an innovative BDFSS is proposed. This strategy is developed through the establishment of a bionic mapping mechanism between fitness fields and light-intensity fields, thereby introducing a novel optimization paradigm.
4) The EDMO's exploration and exploitation abilities were rigorously tested using the CEC2017 and CEC2020 benchmark functions for optimization. The tests clearly demonstrated the algorithm's skill in navigating and utilizing the solution space effectively.
5) EDMO is applied to three engineering design tasks and UAV trajectory planning problems to verify its effectiveness and accuracy in tackling real-world challenges. This highlights the algorithm's applicability in addressing complex engineering issues, particularly in enhancing the performance of autonomous UAV navigation systems.

The structure of this paper is organized as follows: Section 2 reviews the research progress of the DMO algorithm and its applications in UAV path planning, establishing the context for this study. Section 3 presents the original DMO, laying the groundwork for the proposed enhancements. Section 4 delineates three enhancement strategies systematically incorporated into the framework: OLOBL strategy, DQTOS, and BDFSS. These methodologies are designed to address inherent limitations of the baseline algorithm by enhancing convergence robustness and global exploration capacity through synergistic mechanism integration. Section 5 verifies the effectiveness of the proposed algorithm through comprehensive benchmarking, highlighting its superior convergence, robustness, and global search capabilities. Section 6 demonstrates the practicality of EDMO in real-world scenarios, such as engineering design and UAV 3D path planning, confirming its adaptability and optimization potential. Finally, Section 7 concludes the paper by summarizing the findings and offering insights into future research directions.

## 2 Relation works

In recent years, the study of three-dimensional UAV path planning has become a hot research topic, particularly as UAVs find increasingly broad applications across various fields [24, 25]. Effective path planning is essential for ensuring the safe and efficient completion of UAV tasks [26]. Consequently, many optimization algorithms and strategies have been introduced to address this need. Given the NP-hard complexity and the requirement for real-time processing in UAV 3D flight path planning, swarm intelligence optimization algorithms have been widely adopted in this area [23]. This section will discuss the latest developments in the DMO algorithm and the application of heuristic algorithms to tackle the challenges of 3D path planning for UAVs. Table 1 summarizes the recent enhancements in DMO.

Shi et al. developed an adaptive grey wolf optimization algorithm that incorporates a spiral update position method, significantly reducing the average UAV flight time by 22.8%, shortening the algorithm's convergence time, and resulting in smoother UAV flight paths [27]. Zhu et al. enhanced the PSO algorithm by integrating an improved nonlinear dynamic inertia weight, thereby accelerating its convergence speed and improving its fitness function value. They further augmented the algorithm with adaptive speed adjustment, chaotic initialization, and an improved logistic chaos map, which significantly boosted convergence speed, reduced the fitness function value and initialization time, and achieved smoother paths [28]. Hu et al. improved the HBA algorithm by incorporating the Bernoulli shift map, segmental optimal decreasing neighborhood, and horizontal crossover with strategy adaptation, effectively applying it to UAV path planning [29]. Pan and his team proposed a dual-learning strategy golden eagle optimization algorithm, successfully generating efficient and feasible UAV paths for power detection [30]. Jiang et al. developed a diversified group teaching optimization algorithm based on a segmental fitness strategy for navigating a 3D flying environment with multiple obstacles [31]. Zhang et al. introduced a novel fruit fly optimization algorithm that combines phase angle coding with a mutation adaptation mechanism, aimed at optimizing UAV flight paths in complex 3D terrains with various ground defense systems [32]. Finally, Chen et al. presented a flower pollination algorithm enhanced with neighborhood global learning to address UAV path planning challenges [33].

Numerous researchers have extensively utilized MH algorithms for UAV path planning, with some focusing on enhancing evolutionary algorithms and reporting positive results [34]. However, despite the application of the DMO algorithm in various fields, its use in UAV path planning remains relatively limited. To achieve optimal path planning, a more thorough exploration of the two fundamental mechanisms of swarm intelligence algorithms—exploitation and exploration—is essential [35].

In this research, the EDMO algorithm, enhanced by integrating multiple strategies, effectively maintains a balance between exploration and exploitation processes. Experimental results indicate that the performance of the improved method surpasses mainstream algorithms such as CMA-ES [36], LSHADE_cnEpsin [37], LSHADE [38], the original DMO [20], GSA [39], IPO [40], MSMA [41], GWCA [42], OMA [43], CO_DWO [44], and BDMSAO [45]. The enhanced algorithm demonstrates superior results in the context of UAV three-dimensional path planning.

Table 1. Improving DMO in recent years.

| Researcher(s) | Contribution | Reference | Year |
|---|---|---|---|
| Agushaka, J.O. et al. | Proposed ADMO algorithm to address the slow convergence of DMO, introducing behaviors like predation, mound protection, breeding, and group fission to enhance optimization performance. | [46] Journal of Bionic Engineering | 2023 |
| Aldosari F et al. | Improved DMO focusing on constrained engineering design problems and global optimization and data clustering applications. | [47] Symmetry | 2023 |
| Akinola, OA et al. | Combined DMO with Simulated Annealing (SA) to form a hybrid binary variant for feature selection on high dimensional multi-class datasets. | [45] Scientific Reports | 2023 |
| Karthi, M. et al. | Proposed Adaptive Dwarf Mongoose Algorithm for enhanced recognition of overlapped English cursive characters. | [48] SIViP | 2023 |
| Akinola OA et al. | Proposed Binary Dwarf Mongoose Optimizer for solving high-dimensional feature selection problems. | [49] PLOS ONE | 2023 |
| Al-Shourbaji, I et al. | Combined Artificial Ecological Optimization with DMO for feature selection and global optimization problems. | [50] International Journal of Computational Intelligence Systems | 2023 |
| Fu, SW et al. | Improved DMO's search process with novel nonlinear control and exploration strategies, enhancing convergence speed and precision. | [39] Expert Systems with Applications | 2023 |
| Rizk-Allah, RM et al. | Enhanced DMO for identification of unknown parameters in electrical 1-phase transformers. | [51] Neural Computing and Applications | 2023 |
| Al-Shourbaji, I et al. | Proposed AEO-DMOA hybrid method for better balance between exploration and exploitation in search space. | [50] International Journal of Computational Intelligence Systems | 2023 |
| M Abdelrazek et al. | Proposed Chaotic Dwarf Mongoose optimization algorithm for feature selection. | [52] Scientific reports | 2024 |
| Emine BAS | Proposed binary DMO (BinDMO) algorithm to update the binary optimization problem. | [53] Neural Computing and Applications | 2024 |

## 3 The original DMO

The DMO is inspired by the semi-nomadic behavior of dwarf mongooses and represents a method of swarm optimization. In this algorithm, the mongoose swarm is categorized into three functional groups: Alpha, Scout, and Babysitter. The primary role of the Alpha group is foraging and directing the search for food. Role exchanges between the Alpha and Babysitter groups occur when specific criteria are met, facilitating the adaptation of the swarm's strategy. Meanwhile, the Scout group is tasked with locating sleeping mounds, which are essential for maintaining the swarm's equilibrium. The Babysitter group plays a pivotal role in the algorithm's performance, as its size influences the swarm's collective behavior. Additionally, this group is responsible for initiating the exchange of certain parameters, enabling the algorithm to adjust its approach in response to varying problems and conditions [20].

### 3.1 Alpha group

Each member is initialized using Equation 1. Once initialization is complete, the fitness probability value for each mongoose in the population is calculated using Equation 2. This value is then used to select the leader of the Alpha group, denoted as $\alpha$.

$$x_{i,j} = unfrnd(lb, ub, Dim) \tag{1}$$

$$\alpha = \frac{fit_i}{\sum_{i=1}^{N} fit_i} \tag{2}$$

where, $x_{i,j}$ represents the initial position, $unfrnd$ denotes a uniformly distributed random number, $lb$ and $ub$ denote the lower and upper boundaries of the problem, $Dim$ represents the dimension of the decision variable, $fit_i$ is the fitness of the $i^{th}$ individual, and $N$ represents the population size. The number of babysitters is set to $bs$, so the number of individuals in the Alpha group, $n$, is $N - bs$. The foraging path is chosen by $\alpha$ and is influenced by $peep$. The new position of the food source is modeled as shown in Equation 3:

$$x_{i+1} = x_i + phi \times (x_i - x_{rand}) \tag{3}$$

where, $x_{i+1}$ represents the new position of the found food source, $x_i$ represents the current position of the female leader, and $phi$ represents a random number uniformly distributed between [-1,1]. According to literature [20], in this paper, $peep$ is chosen as 2. The initial sleeping mound is set as $\emptyset$, and $x_{rand}$ represents the position of a random individual in the Alpha group. Evaluate the fitness of the alpha group $fit_{i+1}$ for $x_{i+1}$, and calculate the sleeping mound value $sm_i$ according to Equation 4:

$$sm_i = \frac{fit_{i+1} - fit_i}{max(|fit_{i+1}, fit_i|)} \tag{4}$$

The average value $\varphi$ of the sleeping mound and the direction vector $\vec{M}$ determining the movement of the mongoose to the new sleeping mound are calculated according to Equations 5 and 6.

$$\varphi = \frac{\sum_{i=1}^{N} sm_i}{n} \tag{5}$$

$$\vec{M} = \sum_{i=1}^{N} \frac{X_i \times sm_i}{X_i} \tag{6}$$

When the babysitter exchange condition $C \geq L$ is met, members of the Alpha group and Babysitter group will be swapped. The positions of the babysitter group $bs$ are then reinitialized, and their fitness values are recalculated, activating the scouting phase. Here, $C$ represents the time counter, and $L$ represents the babysitter exchange parameter.

### 3.2 Scout group

When the Alpha group has found sufficient food and receives the exchange signal from the babysitters, the Scout group is activated and begins searching for new sleeping mounds. If they explore far enough, they might discover a new sleeping mound. The position $x_{sm}$ of the new sleeping mound is simulated according to Equations 7 and 8.

$$CF = \left(1 - \frac{t}{T}\right)^{2 \times \frac{t}{T}} \tag{7}$$

$$x_{sm} = \begin{cases} x_i - CF \times phi \times rand \times (x_i - \vec{M}), & if\ \varphi_{i+1} > \varphi \\ x_i + CF \times phi \times rand \times (x_i - \vec{M}), & else \end{cases} \tag{8}$$

where, $rand$ represents a random number between 0 and 1, $CF$ represents a parameter representing the mongoose swarm's mobility capability, which linearly decreases with the iteration number. $t$ represents the current iteration number, and $T$ represents the maximum number of iterations.

### 3.3 Babysitter group

The Babysitter group in the algorithm typically comprises secondary swarm members tasked with caring for the young. They rotate frequently to allow the female leader $\alpha$ to lead other members in daily foraging activities. The size of the Babysitter group varies with the overall swarm size, and its impact on the algorithm can be fine-tuned by altering its proportional size within the swarm. Adjusting the Babysitter group's exchange parameter $L$ enables the reconfiguration of scouting and food source data previously gathered by other family members. To ensure a decrease in the Alpha group's average weight in subsequent iterations, the fitness value of the Babysitter group is set to zero, which in turn limits the swarm's movement. This mimics the real-life scenario where the swarm's mobility is constrained due to the necessity of caring for the young.

From this optimization process, it's evident that the algorithm primarily updates its solutions based on interactions and following behavior among individuals, influenced by individual characteristics and position update variables. However, the lack of a mutation strategy means that when the algorithm encounters a local optimum, escaping from the current local search area becomes a significant challenge.

## 4 The proposed EDMO

Considering the aforementioned analysis, we improved the DMO algorithm from three perspectives:

1) An OLOBL strategy was constructed using lens refraction reverse learning and orthogonal experimental design, which enhances the algorithm's exploration ability and addresses the dimensional degradation problem of reverse learning.
2) A Quantum Tunneling Optimization Strategy (QTOS) is proposed, wherein particles are endowed with quantum tunneling capability to penetrate local optima energy barriers. This mechanism significantly enhances global search performance in engineering applications through systematic energy barrier transcendence.
3) A Bio-phototaxis Dynamic Focusing Strategy (BDFS) is proposed, establishing a novel optimization paradigm for the Enhanced Distributed Multi-Objective (EDMO) algorithm through the development of a bionic mapping mechanism between fitness fields and light-intensity gradients.

This section highlights the proposed EDMO algorithm, which incorporates orthogonal learning, quantum-inspired tunneling, and phototactic foraging strategies, supported by adaptive initialization and control mechanisms, to effectively balance exploration and exploitation in complex optimization tasks.

### 4.1 Orthogonal refracted opposition-based learning strategy

In view of the weak ability of MH algorithm to jump out of local optimum, lens imaging opposition-based learning strategy (OLOBL) [54, 55] is introduced to improve the performance of the algorithm, and the optimal solution is sought by generating the reverse position according to the current individual position. This method can effectively expand the search range and increase the chance of finding better solutions. The OLOBL strategy significantly improves the global search ability and the ability to jump out of the local optimum while maintaining the simplicity of the algorithm. The specific principle is as follows:

Suppose there is an individual $P$ in the spatial extent of the interval $[lb, ub]$ with height $h$ and projection $X$ on the $x$-axis. The object $P$ is positioned at a distance $u$ from the convex lens, while its image $P'$ is formed at a distance $v$ on the opposite side. The focal length of the lens, $f$, determines the relationship between $u$, $v$. By imaging with a convex lens placed at point $o$ (which is the midpoint of $[lb, ub]$), $P'$ of height $h'$ can be obtained, and its projection on the $x$-axis is $X'$. Then the imaging principle can be obtained as follows:

$$\frac{\frac{ub+lb}{2} - X}{X' - \frac{ub+lb}{2}} = \frac{h}{h'} \tag{9}$$

where, let $\frac{h}{h'} = k$, and transform the formula to get:

$$X' = \frac{ub+lb}{2} + \frac{ub+lb}{2k} - \frac{X}{k} \tag{10}$$

The scaling factor $k$ is calculated as follows:

$$k = \left(1 + \left(\frac{t}{T}\right)^2\right)^{10} \tag{11}$$

The lens imaging reverse learning strategy explores previously uncovered areas in the solution space by reflecting and scaling solutions, thereby increasing solution diversity and reducing the risk of the algorithm becoming trapped in local optima. Additionally, in the later stages of the algorithm, when the value of $k$ is large, the newly generated solutions become more concentrated around the current optimal solution. This concentration helps the algorithm fine-tune these solutions more precisely, accelerating convergence to the global optimum or near-global optimum solutions.

Orthogonal experimental design (OED) can identify the optimal experimental combination for multiple factors and levels through a reduced number of tests [56]. For instance, in an experiment with 2 levels and 7 factors, a full factorial test would require $2^7 = 128$ tests to find the optimal combination. In contrast, using the orthogonal table $L_8(2^7)$ as shown in Equation (12), allows for the discovery of optimal or near-optimal combinations with only 8 tests, significantly enhancing experimental efficiency. However, due to the nature of orthogonal experimental design, it cannot guarantee that the solutions in the orthogonal table include the true optimal solution [56]. Therefore, it is generally necessary to perform factor analysis to identify the theoretically optimal combination and then compare this with all combinations in the orthogonal table to determine the final optimal solution. Thus, for the experiment with 2 levels and 7 factors, one must first obtain 8 candidate optimal solutions from the orthogonal table $L_8(2^7)$, conduct factor analysis to identify a theoretically optimal combination, and then evaluate the 9 combinations to determine the overall optimal solution for the experiment.

$$L_8(2^7) = \begin{bmatrix} 1 & 1 & 1 & 1 & 1 & 1 & 1 \\ 1 & 1 & 1 & 2 & 2 & 2 & 2 \\ 1 & 2 & 2 & 1 & 1 & 2 & 2 \\ 1 & 2 & 2 & 1 & 1 & 1 & 1 \\ 2 & 1 & 2 & 1 & 2 & 1 & 2 \\ 2 & 1 & 2 & 2 & 1 & 2 & 1 \\ 2 & 2 & 1 & 1 & 2 & 2 & 1 \\ 2 & 2 & 1 & 2 & 1 & 1 & 2 \end{bmatrix} \tag{12}$$

To enhance the DMO algorithm's ability to escape local optima, this paper proposes a OLOBL strategy, which is applied to the leader individual to generate new candidate individuals.

OLOBL is a strategy developed by integrating orthogonal experimental design (OED) and lens opposition-based learning techniques. The optimal solution executes the OLOBL strategy, enabling jumps to more promising search areas, thereby enhancing population diversity and reducing the likelihood of the algorithm becoming trapped in local optima. However, the study referenced in [57] indicates that an individual's opposite solution is only superior to the current solution in certain dimensions. To address this issue, an orthogonal reflection opposite learning strategy is proposed, which fully explores each dimensional component of both the current and opposite solutions, combining their advantageous dimensions to generate a partial reflection opposite solution.

The OLOBL strategy is integrated into the DMO algorithm, where the optimization problem's dimension $D$ corresponds to the factors in the orthogonal experimental design, and the individual and its reflection opposite solution represent the two levels of this design. The detailed process for constructing a partial reflection opposite solution is as follows: an orthogonal experiment with 2 levels and $D$ factors is designed for the current solution and its reflection opposite solution, generating $M$ partial reflection opposite solutions, where $M$ is calculated according to Equation (13). Specifically, when generating partial opposite solutions based on the orthogonal table, if an element in the table is 1, the corresponding dimension in the trial solution is set to the value of the current solution; if the element is 2, the corresponding dimension takes the value of the reflection opposite solution.

$$M = 2^{log_2(D+1)} \tag{13}$$

According to the characteristics of orthogonal experimental design, all elements in the first row of the orthogonal table are 1, indicating that the first trial solution is identical to the original individual and does not require evaluation. The remaining $(M-1)$ trial solutions are combinations of the advantageous dimensions of the current individual and its reflection opposite individual, referred to as partial reflection opposite solutions, which do need evaluation. When using orthogonal experimental design, factor analysis is necessary to identify a theoretically optimal combination that may not exist in the orthogonal table, which also requires evaluation. Therefore, executing the OLOBL strategy necessitates $M$ function evaluations. During the evolutionary iterations, the OLOBL strategy is applied only to the leader, and the superior individual is selected from the leader and its orthogonal reflection opposite solutions to enter the next generation. This approach effectively enhances the global exploration ability of the algorithm, reduces the number of function evaluations, and improves the overall performance of the algorithm.

In the orthogonal reflection opposite learning strategy, a reflection opposite learning approach based on the lens imaging principle is employed to enhance exploration of the opposite solution space, significantly reducing the likelihood of the algorithm becoming trapped in local optima. The orthogonal experimental design is utilized to construct several partial opposite solutions by taking reflection opposite values in certain dimensions, thoroughly exploring and preserving the advantageous dimensional information of both the current individual and the reflection opposite individual.

**4.2 Dynamic Quantum Tunneling Optimization Strategy**

Traditional swarm intelligence algorithms often fail to converge to the global optimum when dealing with high-dimensional complex optimization problems, as individuals tend to become trapped in local optima. Inspired by the phenomenon of particle tunneling through potential barriers in quantum mechanics, this section proposes a perturbation strategy based on the DQTOS. As shown in Figure 1, This strategy endows individuals with non-classical probabilistic transition capabilities, enabling them to penetrate the "energy barriers" formed by local optima. Consequently, the global exploration capability of the algorithm is enhanced in engineering scenarios such as three-dimensional path planning for unmanned aerial vehicles.

Let the position of the best individual in the current population be denoted as $X_{\text{best}} \in \mathbb{R}^d$, with its fitness value represented as $f_{\text{best}}$. The quantum potential energy function of individual $X_i$ is defined as follows:

$$V(X_i) = \frac{|fit_i - f_{\text{best}}|}{\max(f) - \min(f)} \cdot V_0 \tag{14}$$

Where, $\max(f)$ and $\min(f)$ denote the maximum and minimum fitness values of the current population, respectively; $V_0 \in \mathbb{R}^+$: the baseline barrier height, with an initial value set at 0.5.

The function maps the fitness differences to the quantum barrier heights. When the fitness of an individual approaches the current optimum, the corresponding potential energy decreases. Conversely, a high potential barrier that impedes the movement of the individual is formed.

In accordance with the Wentzel-Kramers-Brillouin (WKB) approximation theory in quantum mechanics, the probability of a particle tunneling through a barrier is related to the shape of the barrier and the energy of the particle. To simplify the computation, an equivalent tunneling probability formula is designed.

$$P_t(X_i) = \exp\left(-\frac{2\sqrt{2m}}{\hbar}\int_a^b \sqrt{V(x) - E}\,dx\right) \tag{15}$$

where, $E_k = \frac{1}{N}\sum_{i=1}^{N} f(X_i)$: the average kinetic energy of the population, which reflects the activity level of the group; $\gamma \in [0.05, 0.2]$ : the tunneling adjustment coefficient (value is 0.1 );When $V(X_i) \leq E_k, P_t$ is set to 1 to ensure free penetration in regions with low potential barriers.

The probability model possesses a distinct physical significance. When the potential energy at an individual's location is lower than the population's average kinetic energy ($V(X_i)/E_k < 1$), the tunneling probability increases exponentially with the decrease of the ratio. Conversely, it rapidly diminishes when the potential energy exceeds the average kinetic energy.

If a random number $rand < P_t(X_i)$, quantum tunneling behavior is triggered. The position of the individual is updated according to the following rules:

$$X_{\text{new}} = X_i + \beta \cdot T \cdot \Delta \tag{16}$$

The parameters in the formula are defined as follows: $\beta \sim \mathcal{N}(0,1)$ denotes a random directional vector following the standard normal distribution. $T(t) = \frac{1}{1+\exp(t/\tau)}$ represents the time-varying temperature coefficient. The attenuation constant $\tau = 0.1T_{max}$ controls the rate at which the exploration intensity decays over time; $\Delta(t) = \frac{ub-lb}{2} \cdot \left(1 - \frac{t}{T_{max}}\right)^2$ is the dynamic step size factor. It allows extensive exploration in the initial phase and focuses on local fine-tuning in the later stages.

The update formula achieves intelligent perturbation through the coupling of three terms: $\beta$ provides a random search direction, $T(t)$ balances the exploration and exploitation phases, and $\Delta(t)$ ensures the adaptive contraction of the search step size.

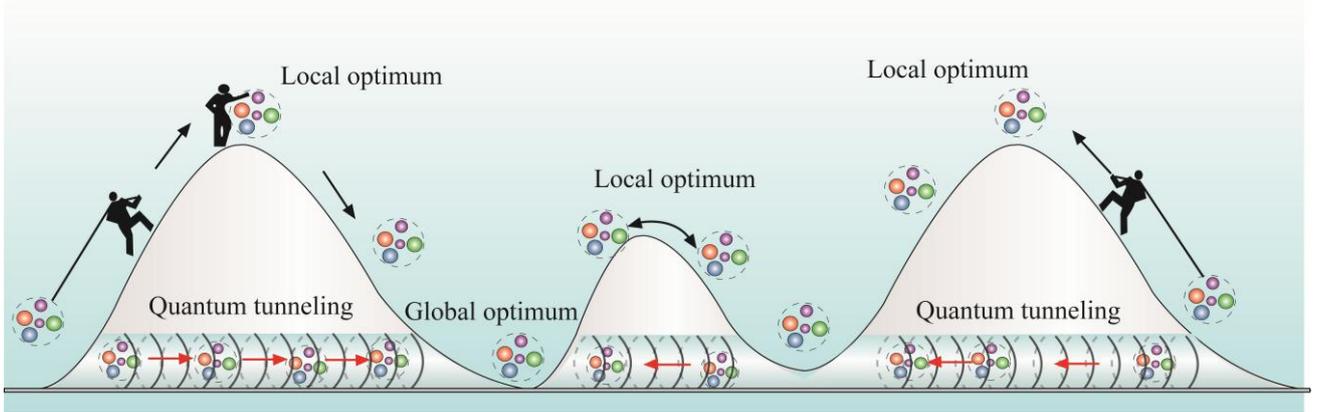

Figure 1. Dynamic Quantum Tunneling Optimization Strategy.

### 4.3 Bio-phototactic Dynamic Focusing Search Strategy

The application of DMO in complex optimization problems is increasingly widespread. However, traditional methods still suffer from premature convergence and low search efficiency when dealing with high-dimensional non-convex problems. It is worth noting that the phototactic behavior of microbial populations in nature is characterized by self-organization, adaptability and efficient resource foraging, as shown in Figure 2. Inspired by this, this section proposes an innovative BDFSS. By establishing a bionic mapping relationship between the fitness field and the light intensity field, it provides a new optimization paradigm for the EDMO algorithm.

To effectively transform biological behavior into an optimization algorithm, three key issues must be addressed: (1) how to quantitatively define the virtual light intensity field; (2) how to mathematically model the individual response mechanism; and (3) how to integrate the collective behavior with the existing algorithm framework. The following sections will elaborate on the construction of these components.

Define the light intensity field function in the solution space:

$$I(X,t) = \frac{fit_i - f_{\text{worst}}(t)}{f_{\text{best}}(t) - f_{\text{worst}}(t)} \cdot I_{\text{max}} \tag{17}$$

where, $f_{\text{best}}(t)$ represent the best fitness value of the population at generation $t$; $f_{\text{worst}}(t)$ denote the worst fitness value of the population at generation $t$; $I_{\text{max}}$ represent the maximum light intensity reference value.

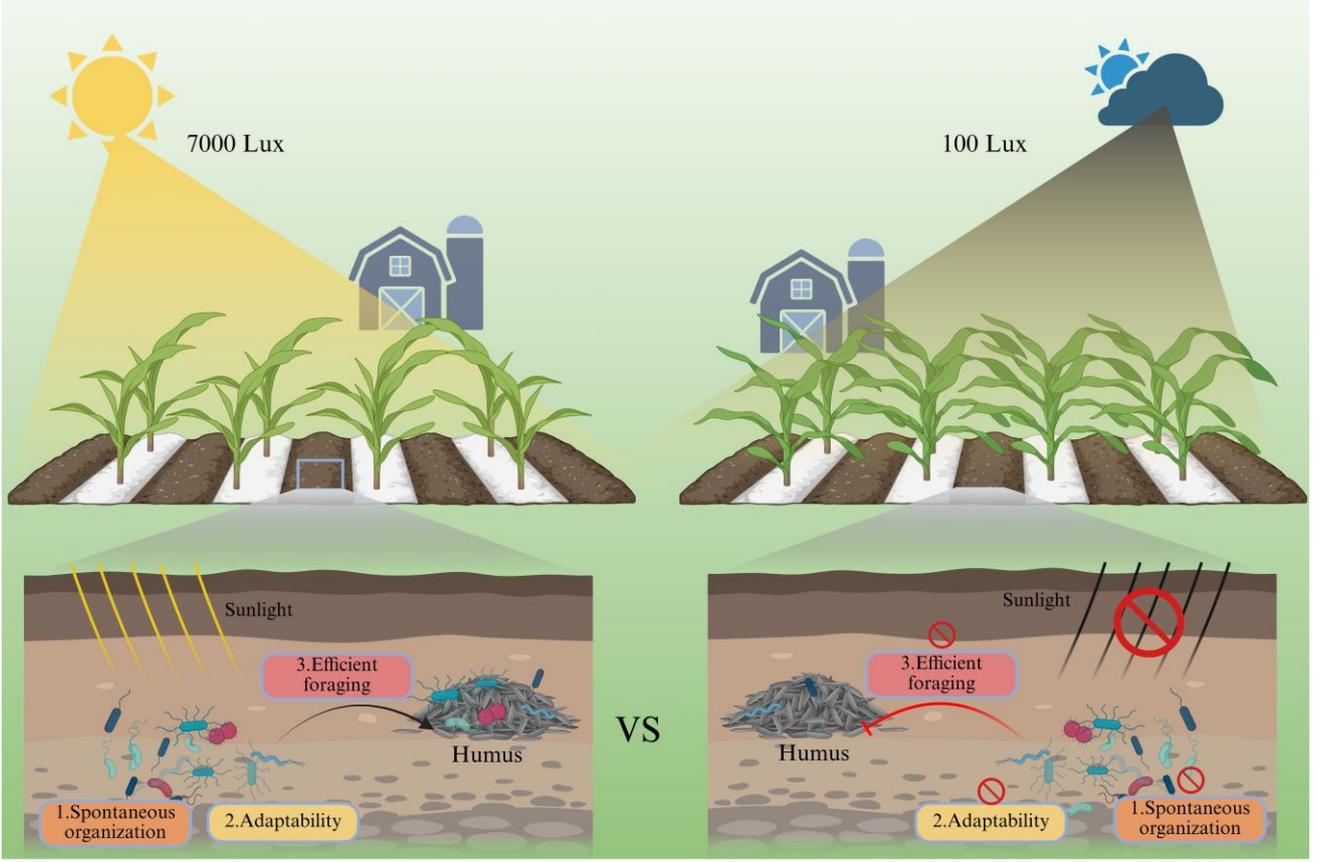

Figure 2. Bio-phototactic Dynamic Focusing Inspiration Diagram.

This mapping transforms the fitness landscape into a virtual light intensity distribution, such that the current optimal solution corresponds to the peak light intensity $(I = 1)$, the worst solution corresponds to the lowest light intensity $(I = 0)$, and the remaining individuals are distributed within the interval [0,1] in proportion to their fitness values. The establishment of the light intensity field provides a foundation for simulating the phototactic behavior of biological organisms.

Define the dynamic response coefficient of an individual:

$$\alpha_i(t) = \alpha_{\min} + (\alpha_{\max} - \alpha_{\min}) \cdot \frac{I(X_i, t)}{I_{\max}} \tag{18}$$

where, $\alpha_{\min} = 0.1$ is the minimum sensitivity, ensuring basic exploration ability, and $\alpha_{\min} = 0.9$ is the maximum sensitivity, preventing excessive exploitation.

The light intensity gradient is calculated using the central difference method:

$$\frac{\partial I}{\partial x_k} \approx \frac{I(X + \delta e_k) - I(X - \delta e_k)}{2\delta} \tag{19}$$

where $\delta = 0.01(ub_k - lb_k)$ is the adaptive step size, and $e_k$ is the unit vector in the $k$-th dimension.

Although the behavioral model of a single individual has been established, the strength of swarm intelligence lies in the collaborative interaction among individuals, which necessitates the design of an appropriate information exchange mechanism.

Hybrid-Driven Position Update:

$$x_{\text{sm}} = \begin{cases} x_i - CF \cdot \varphi \otimes \left(\Delta X_{\text{social}} + \xi \cdot \alpha_i(t)\right), & \varphi_{i+1} > \varphi \\ x_i + CF \cdot \varphi \otimes \left(\Delta X_{\text{photo}} + \xi \cdot \frac{\partial I}{\partial x_k}\right), & \text{otherwise} \end{cases} \tag{20}$$

where $\xi$ present the hybrid perturbation factor, $\xi = 0.5 \cdot \left(1 - \frac{t}{T}\right)$; $\otimes$ denotes the Hadamard product, which represents element-wise multiplication.

The light intensity gradient guides the population to rapidly explore potential optimal regions, avoiding blind search. Under high sensitivity, individuals make fine adjustments to their positions, thereby enhancing the convergence accuracy. This integrated strategy offers a more efficient solution for complex engineering optimization problems and holds broad application prospects, particularly in the fields of autonomous navigation of unmanned aerial vehicles and intelligent manufacturing scheduling.

### 4.4 EDMO algorithm description

Within the EDMO framework, the positional update of each individual is governed not only by its assigned role but also through coordinated interactions with other group members. This methodology is explicitly designed to augment global search performance. Post-positional updates, the algorithm integrates a dual-strategy mechanism: a DQTOS inspired by quantum tunneling phenomena to overcome local optima energy barriers, and an Innovative BDFS derived from the self-organizing and adaptive phototactic behaviors observed in microbial populations. These biologically and physically inspired strategies synergistically enhance individual agents' ability to evade local optima traps while facilitating the exploration of advantageous positions, thereby optimizing fitness evaluation outcomes. The EDMO workflow, as illustrated in Figure 3, is executed through the following sequential phases:

a）The procedure commences with stochastic initialization of the population. A population of candidate solutions is stochastically generated through randomized parameter assignments. This initial solution set serves as the foundational population for subsequent evolutionary operations. The initialization mechanism employs uniform probability distributions across all design variables. Each individual in the population vector is instantiated with independent random values within predetermined parameter bounds.

b）Through the implementation of OLOBL strategy, the comprehensive position is generated, and then the fitness evaluation is carried out. The fitness evaluation stage involves the strict quantification of the objective function value of each individual.

c）The elite individuals are retained, and the individuals with the highest fitness are selected for retention to maintain the excellent solution in the population, calculate the light intensity field, calculate the light intensity of each individual, and prepare for the later update sensitivity.

d）Then update the sensitivity $\alpha$: adjust the sensitivity parameters in the algorithm to affect the individual's response to light intensity, and then update the hybrid drive position to update the individual's position according to the light intensity and sensitivity.

e）If a probability $P_t$ of the individual is greater than the random number rand, the position jump is performed, otherwise the current position is maintained.

f）Judge $C \geq L$? If a condition $C$ is greater than $L$, the group position exchange is performed, otherwise the sleep heap value is updated.

g）To determine $t \geq T_{Max}$? If the current number of iterations $t$ reaches the maximum number of iterations $T_{\max}$, the optimal solution is output and the algorithm is ended.

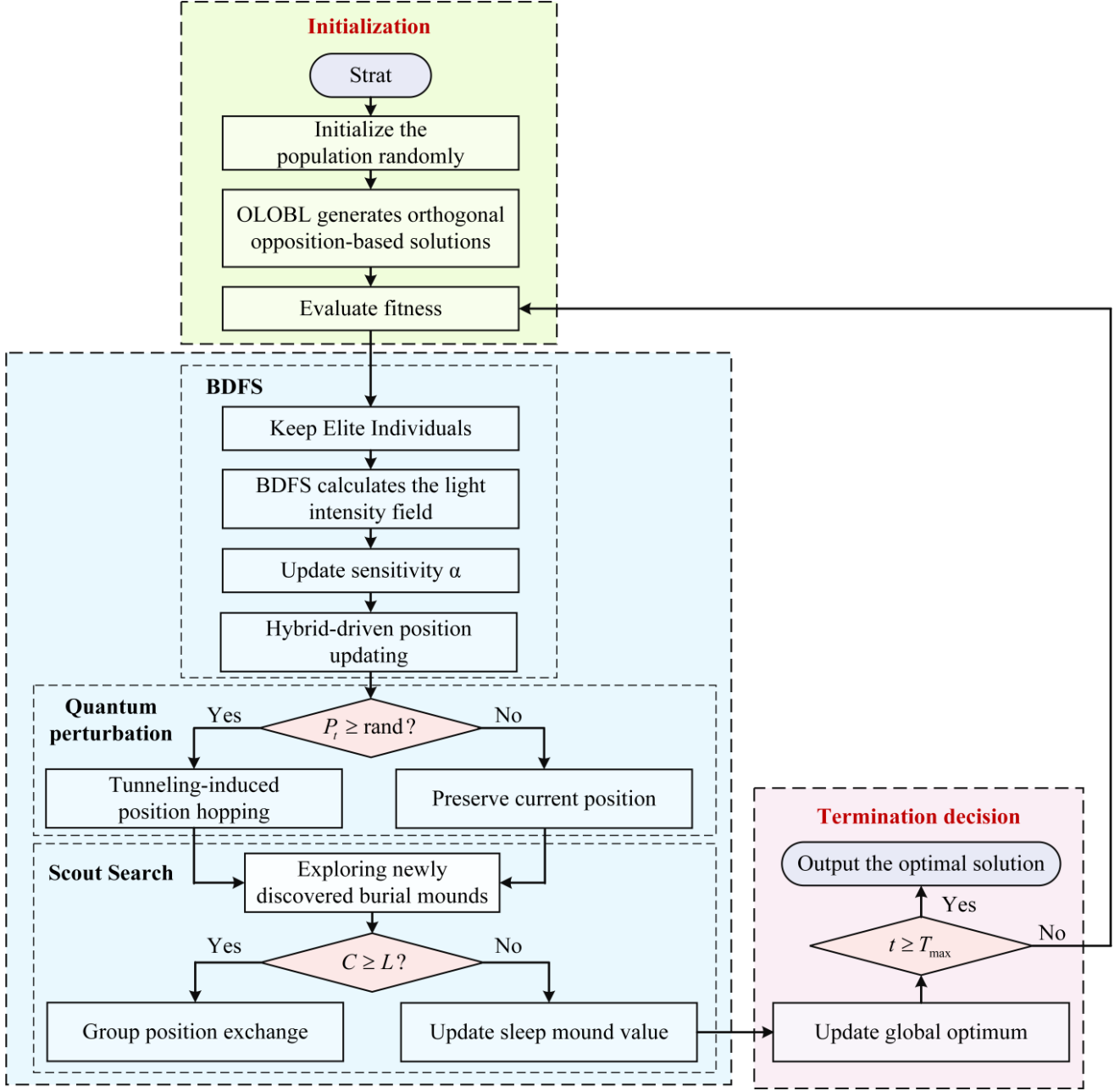

Figure 3. Flow chart of EDMO.

### 4.5 The computational complexity analysis

The DMO algorithm has a time complexity of $O(N \times d \times M)$, where N is the population size, $d$ is the dimension, and $M$ is the maximum number of iterations. The time complexity analysis of the EDMO algorithm is outlined below:

1) Introducing the OLOBL strategy for population initialization has a time complexity of $O(N \times d \times M)$. Therefore, the overall time complexity considering the OLOBL strategy is $O(N \times d \times M + N \times d \times M)$, which simplifies to $O(N \times d \times M)$.

2) Despite the introduction of additional quantum tunneling computations by the DQTOS, its overall time complexity remains on the same order of magnitude as that of the original EDMO. Specifically, the calculation of potential energy, which involves computing fitness differences for each individual, has a time complexity of $O(N)$. The evaluation of tunneling probability, although involving exponential operations, can be optimized to $O(1)$ per individual through the use of a lookup table. Furthermore, the random number generation and vector operations required for position updates are $O(d)$ per individual. Therefore, the additional computational cost per iteration of QTPS is $O(N \times d)$, which is negligible when compared to the $O(N \times d \times M)$ complexity of the EDMO framework.

3) Calculating the light intensity value for individual has a complexity of $O(N \times d)$. Determining $\alpha_I$ based on these light intensity values also has a complexity of $O(N \times d)$. The hybrid update operation involves a combination of operations with a complexity of $O(N \times d)$. Therefore, the complexity per generation is $O(N \times d)$, without introducing any additional complexity. which simplifies to $O(N \times d \times M)$.

In summary, the time complexity of the EDMO algorithm remains $O(N \times d \times M)$, same as the standard DMO algorithm. This indicates that compared to the standard DMO, the time complexity of EDMO does not increase.

## 5 Algorithm performance testing and analysis

The simulations in this study were conducted on a system running Windows 10 64-bit, powered by an AMD Ryzen 74800H CPU with a 2.30GHz base frequency, and equipped with 16GB RAM. The algorithms, including the QANA, SMO, IAO, CMA-ES, LSHADE_cnEpsin, LSHADE, the original DMO, GSA, IPO, MSMA, GWCA, OMA, CO_DWO, and BDMSAO, were implemented in MATLAB 2023b.

### 5.1 Parameter selection experiment

In EDMO, the parameter *peep* simulates the behavior of dwarf mongooses and guides the algorithm in generating candidate solution locations. Different values of *peep* the CEC2017 test suite is used to test the performance of EDMO with different *peep,* the CEC2017 test suite is employed to assess the performance of EDMO with $peep = [1,2,3,4]$, and the results are recorded in Table 2. This table presents the best fitness mean (Ave) and standard deviation (Std) for EDMO with each *peep* value over 30 repeated experiments. The algorithm with the best mean value for each group of test functions is highlighted in bold. The optimal values for repeated trials are shown in bold for a test dimension $d = 30$. From the perspective of optimization accuracy, the statistical optimal value when $peep = 2$ slightly outperforms the others, achieving the most optimal or near-optimal performance on the majority of test functions, as indicated by the Ave and Std values. Therefore, $peep = 2$ is selected as the EDMO parameter in this paper.

Table 2. The results of the EDMO for different values of *peep*.

| ID | Metric | EDMO (*peep*=1) | EDMO (*peep*=2) | EDMO (*peep*=3) | EDMO (*peep*=4) |
| --- | --- | --- | --- | --- | --- |
| CEC2017-F1 | Ave | 6.0416E+05 | **5.3879E+05** | 5.8982E+05 | 6.7153E+05 |
| | Std | 1.7020E+05 | 1.5730E+05 | **1.5498E+05** | 1.7479E+05 |
| CEC2017-F3 | Ave | **1.1481E+03** | 1.3258E+03 | 1.6863E+03 | 2.0332E+03 |
| | Std | 6.8518E+02 | **5.3829E+02** | 6.7442E+02 | 9.1530E+02 |
| CEC2017-F4 | Ave | **4.7044E+02** | 4.7610E+02 | 4.8046E+02 | 4.7171E+02 |
| | Std | 3.4777E+01 | 2.9795E+01 | 2.9467E+01 | **2.8840E+01** |
| CEC2017-F5 | Ave | 6.5720E+02 | **6.4183E+02** | 6.4572E+02 | 6.4570E+02 |
| | Std | 3.0438E+01 | 3.4181E+01 | 3.6189E+01 | **2.9269E+01** |
| CEC2017-F6 | Ave | 6.3120E+02 | **6.2796E+02** | 6.2872E+02 | 6.2818E+02 |
| | Std | 1.2840E+01 | 1.4212E+01 | **9.4197E+00** | 1.2333E+01 |
| CEC2017-F7 | Ave | 8.6080E+02 | **8.4167E+02** | 8.5454E+02 | 8.5593E+02 |
| | Std | 4.0115E+01 | **2.5059E+01** | 3.8844E+01 | 2.8323E+01 |
| CEC2017-F8 | Ave | 9.3161E+02 | **9.3124E+02** | 9.3463E+02 | 9.3257E+02 |
| | Std | 2.9780E+01 | 2.9692E+01 | 2.9528E+01 | **2.2160E+01** |
| CEC2017-F9 | Ave | 3.5123E+03 | **3.2547E+03** | 3.3091E+03 | 3.4356E+03 |
| | Std | 1.4127E+03 | 1.1761E+03 | **1.0863E+03** | 1.4818E+03 |
| CEC2017-F10 | Ave | 4.3866E+03 | 4.2763E+03 | 4.2470E+03 | **4.2117E+03** |
| | Std | **4.9452E+02** | 6.7360E+02 | 6.3765E+02 | 5.6655E+02 |
| CEC2017-F11 | Ave | 1.2439E+03 | 1.2367E+03 | **1.2261E+03** | 1.2272E+03 |

| ID | Metric | EDMO (*peep*=1) | EDMO (*peep*=2) | EDMO (*peep*=3) | EDMO (*peep*=4) |
|---|---|---|---|---|---|
| | Std | 3.7494E+01 | **3.4063E+01** | 3.7584E+01 | 3.7835E+01 |
| CEC2017-F12 | Ave | **6.3960E+06** | 8.5643E+06 | 7.8956E+06 | 9.0055E+06 |
| | Std | 5.7294E+06 | 6.3253E+06 | **5.2547E+06** | 1.0031E+07 |
| CEC2017-F13 | Ave | 3.1146E+05 | 3.1641E+05 | 4.1198E+05 | **2.6513E+05** |
| | Std | 3.4258E+05 | 2.6762E+05 | 3.9898E+05 | **2.1444E+05** |
| CEC2017-F14 | Ave | 4.3355E+04 | 5.9568E+04 | **4.1488E+04** | 4.1763E+04 |
| | Std | **3.9265E+04** | 1.0239E+05 | 4.4555E+04 | 4.0248E+04 |
| CEC2017-F15 | Ave | **4.7113E+03** | 4.6954E+04 | 5.7096E+04 | 5.3697E+04 |
| | Std | **3.4014E+04** | 4.8418E+04 | 5.9096E+04 | 4.2823E+04 |
| CEC2017-F16 | Ave | 2.7074E+03 | **2.6226E+03** | 2.6582E+03 | 2.6315E+03 |
| | Std | 2.4969E+02 | **2.4759E+02** | 2.8134E+02 | 3.1348E+02 |
| CEC2017-F17 | Ave | 2.2557E+03 | **2.2339E+03** | 2.2562E+03 | 2.2480E+03 |
| | Std | 2.0654E+02 | **1.8187E+02** | 2.6848E+02 | 1.9615E+02 |
| CEC2017-F18 | Ave | **4.2497E+05** | 5.2478E+05 | 5.1972E+05 | 5.7852E+05 |
| | Std | 4.1379E+05 | 4.3470E+05 | **3.0190E+05** | 5.1690E+05 |
| CEC2017-F19 | Ave | **1.5226E+04** | 1.6878E+04 | 1.8622E+04 | 1.9940E+04 |
| | Std | **8.7194E+03** | 9.9952E+03 | 1.1730E+04 | 1.1940E+04 |
| CEC2017-F20 | Ave | 2.4691E+03 | **2.4512E+03** | 2.5196E+03 | 2.4638E+03 |
| | Std | 1.7049E+02 | 1.4627E+02 | 1.6370E+02 | **1.4168E+02** |
| CEC2017-F21 | Ave | 2.4283E+03 | 2.4356E+03 | 2.4345E+03 | **2.4265E+03** |
| | Std | 4.9597E+01 | 3.7153E+01 | 3.3648E+01 | **2.6872E+01** |
| CEC2017-F22 | Ave | 2.4390E+03 | 2.3188E+03 | 2.3184E+03 | **2.3123E+03** |
| | Std | 6.5221E+02 | 1.6981E+00 | 1.6321E+00 | **1.3851E+00** |
| CEC2017-F23 | Ave | 2.8447E+03 | 2.8088E+03 | **2.7972E+03** | 2.8274E+03 |
| | Std | 5.1929E+01 | 5.6782E+01 | 4.6518E+01 | **4.3869E+01** |
| CEC2017-F24 | Ave | 3.0239E+03 | 3.0145E+03 | **3.0044E+03** | 3.0940E+03 |
| | Std | 5.3159E+01 | 6.5191E+01 | 5.4588E+01 | **3.9433E+01** |
| CEC2017-F25 | Ave | 2.8914E+03 | **2.8868E+03** | 2.8941E+03 | 2.8991E+03 |
| | Std | 1.2626E+01 | **9.5593E+00** | 2.0633E+01 | 1.3797E+01 |
| CEC2017-F26 | Ave | 3.6746E+03 | 4.3336E+03 | **3.5117E+03** | 3.7711E+03 |
| | Std | 1.1854E+03 | 1.3716E+03 | **1.1136E+03** | 1.2524E+03 |
| CEC2017-F27 | Ave | **3.1941E+03** | 3.2060E+03 | 3.2087E+03 | 3.2094E+03 |
| | Std | 2.0074E+01 | 9.8120E+00 | 2.2345E-04 | **1.7231E-04** |
| CEC2017-F28 | Ave | **3.2516E+03** | 3.2558E+03 | 3.2698E+03 | 3.2938E+03 |
| | Std | 3.1195E+01 | 3.5170E+01 | 4.0268E+01 | **1.3752E+01** |
| CEC2017-F29 | Ave | **3.7566E+03** | 3.7588E+03 | 3.7656E+03 | 3.7591E+03 |
| | Std | 2.5376E+02 | 2.7064E+02 | **1.6513E+02** | 2.1841E+02 |
| CEC2017-F30 | Ave | 8.7178E+04 | **7.9899E+04** | 8.0042E+04 | 9.0696E+04 |
| | Std | 1.2248E+05 | **1.1071E+05** | 1.1558E+05 | 1.2748E+05 |

## 5.2 Test functions and parameter settings

To validate the effectiveness of the proposed EDMO, it is tested using the CEC2017 [58] and CEC2020 [59] test suites. The CEC series of functions comprises a mixture of various basic test functions. These functions are commonly used as benchmarks for comparing the performance of different optimization algorithms and are widely applied to simulate the complexity of real-world problems.

In the experiments, the population size $N$ of the algorithm was set to 50, and the maximum number of iterations

$M$ was set to 500. Each set of experiments was run independently 30 times, and the best fitness value from each run was recorded. To highlight the superiority of the algorithm proposed in this study, it is compared with 14 other competitive algorithms: QANA [60], SMO [61], IAO [62], CMA-ES [36], LSHADE_cnEpsin [37], LSHADE [38], the original DMO [20], GSA [39], IPO [40], MSMA [41], GWCA [42], OMA [43], CO_DWO [44], and BDMSAO [45]. The details of the algorithm parameters are presented in Table 3.

Table 3. Algorithms parameter settings.

| Algorithm | Parameter | Value | Number of evaluations |
|---|---|---|---|
| QANA | k, K, K2 | 10, 9, 50 | 25000 |
| SMO | k, λ, μ, θ, ψ | 10, 20, 0.5, [0.1,8], [0.1,8] | 25000 |
| IAO | $\beta, \varepsilon$ | [0,1], [0,1] | 25000 |
| DMO | peep | 2 | 25000 |
| CMA_ES | ωi | [2.0794, 1.3863, 0.9808, 0.6931, 0.4700, 0.2877, 0.1335] | 25000 |
| LSHADE_cnEpsin | P,|A|,H | 0.11,1.4,5 | 25000 |
| LSHADE | P,|A|,H | 0.11,1.4,5 | 25000 |
| EDMO | peep | 2 | 25000 |
| GSA | α, G0 | 20, 100 | 25000 |
| IPO | k1,k2 | 0.7184,2.7613 | 25000 |
| MSMA | z,E,N | 0.03,100,10 | 25000 |
| GWCA | P,Q,T,M | 9,6,8.3,3 | 25000 |
| OMA | $m^r$ | [0,1] | 25000 |
| CO_DWO | peep | 2 | 25000 |
| BDMSAO | peep, T0, α, μ | 2, 0.1, 0.99, 0.5 | 25000 |

### 5.3 Ablation experiment

To further elucidate the effects of the OLOBL strategy, DQTOS, and BDFS on DMO individually, this paper conducts ablation comparison experiments. The algorithms employing the OLOBL improvement strategy, DQTOS, and BDFS are referred to as LDMO, DDMO, and BDMO, respectively. Specifically, DQTOS in DDMO combines the quantum tunneling ability of particles described in Figure 1 to penetrate the local optimal energy barrier and dynamically break through the local optimal value throughout the optimization process. The OLOBL strategy in LDMO, illustrated through the lens imaging principle, enhances global exploration by systematically searching the solution space, while the BDFS in BDMO promotes resilience and diversity in the search, particularly in complex and high-dimensional spaces. These algorithms are tested using the CEC2017 function test set. The number of runs, population size, test dimensions, and maximum iterations remain consistent with Section 5.2. Repeated experiments are conducted for all the algorithms mentioned above. To quantitatively assess the importance of the omitted strategy, an ablation factor γ is defined, and its calculation formula is as follows:

$$\gamma = \frac{f_{EDMO}}{f_{EDMO_i}} \quad (21)$$

where, $f_{EDMO_i}$ and $f_{EDMO}$ represent the average best fitness values of the partially improved DMO and EDMO, respectively. A smaller γ value indicates poorer optimization performance of the control algorithm, suggesting that the missing strategy has a more significant role in enhancing the performance of the original algorithm. The results of the ablation experiments are presented in Table 4. The γ values of LDMO, ADMO, and SDMO are generally greater than those of DMO, demonstrating that combining any two strategies can enhance DMO's optimization performance.

Table 4. Comparison of ablation results.

| Function | LDMO | DDMO | BDMO | EDMO | Function | LDMO | DDMO | BDMO | EDMO |
|---|---|---|---|---|---|---|---|---|---|
| $f_1$ | 0.00 | 0.31 | 0.72 | **0.03** | $f_{17}$ | 0.85 | 0.96 | 0.97 | **0.88** |
| $f_3$ | 0.52 | 0.51 | 0.92 | **0.32** | $f_{18}$ | 0.02 | 0.78 | 0.78 | **0.32** |
| $f_4$ | **0.82** | 0.83 | 0.87 | 0.87 | $f_{19}$ | 0.39 | 0.54 | 0.78 | **0.48** |
| $f_5$ | 0.87 | 0.93 | 0.93 | **0.82** | $f_{20}$ | 0.88 | 0.96 | 0.95 | **0.92** |
| $f_6$ | 0.98 | **0.97** | 0.98 | 1.00 | $f_{21}$ | 0.99 | 0.97 | 0.98 | **0.95** |
| $f_7$ | 0.89 | 0.92 | 0.88 | **0.87** | $f_{22}$ | **0.57** | 0.68 | 1.06 | 0.77 |
| $f_8$ | 0.96 | 0.96 | 0.97 | **0.91** | $f_{23}$ | **0.97** | 0.98 | **0.97** | 0.97 |
| $f_9$ | **0.34** | 0.50 | 0.49 | 0.68 | $f_{24}$ | 1.00 | **0.97** | 1.02 | 0.98 |
| $f_{10}$ | 0.55 | 0.61 | 0.52 | **0.43** | $f_{25}$ | **0.98** | 0.98 | 0.98 | 0.99 |
| $f_{11}$ | 1.15 | 1.13 | **1.12** | 1.12 | $f_{26}$ | 0.79 | 0.81 | 0.76 | **0.72** |
| $f_{12}$ | **0.38** | 0.65 | 0.45 | 0.42 | $f_{27}$ | **0.94** | 0.99 | 0.99 | **0.98** |
| $f_{13}$ | **0.32** | 0.42 | 0.43 | 0.56 | $f_{28}$ | **0.93** | 0.95 | 0.98 | 0.99 |
| $f_{14}$ | 2.14 | 8.33 | 0.12 | **0.05** | $f_{29}$ | **0.84** | 0.92 | 0.89 | 0.91 |
| $f_{15}$ | 0.23 | **0.22** | **0.22** | 0.24 | $f_{30}$ | 0.07 | 0.44 | **0.22** | 0.44 |
| $f_{16}$ | 0.83 | 0.87 | 0.87 | **0.79** | — | — | — | — | — |

The strategies proposed in this study demonstrate clear advantages across unimodal, multimodal, and hybrid functions. However, the slime mold foraging strategy shows relatively limited improvement on functions $f_1$ and $f_3$. This is primarily because the slime mold foraging strategy is a globally-focused improvement method. When dealing with test functions that demand detailed local searches, this global approach may not be as effective as other strategies that are more specialized in local optimization. Functions $f_1$ and $f_3$ primarily assess an algorithm's fine-tuning capabilities. Such a global approach is less effective for tasks requiring detailed local search but excels in expansive search spaces, enhancing exploration and resilience in problems demanding broader exploration. For function $f_{14}$, the $\gamma$ value of LDMO is considerably higher than for the other algorithms, indicating that the OLOBL improvement strategy is less suited to this function. However, this does not diminish its overall contribution, as it significantly improves performance on functions requiring robust global optimization. Overall, the performance gains from each strategy on DMO do not differ significantly, and the overall improvement of the algorithm is largely dependent on the collaboration among the different strategies.

## 5.4 Comparative analysis of EDMO and other algorithm

The CEC series of functions effectively simulate the complexity of real-world problems, providing valuable insights for developing new algorithms. To maintain parameter consistency, the number of runs, population size, test dimensions, and maximum iterations in the experiments were kept in line with Section 5.2. Specifically, 30 independent runs were conducted, and the best fitness values were recorded for each experiment. Table 5, Table 6 and Table 7 present the average best fitness (Ave) and standard deviation (Std) for QANA, SMO, IAO, CMA-ES, LSHADE_cnEpsin, LSHADE, the original DMO, GSA, IPO, MSMA, GWCA, OMA, CO_DWO, BDMSAO, and EDMO across 30 repeated experiments. The algorithm with the best mean value for each set of test functions is highlighted in bold.

Table 5. Test results for CEC2017 (d=30).

| ID | Metric | QANA | SMO | IAO | CMA_ES | LSHADE | LSHADE_cnEpSin | GSA | IPO | MSMA | GWCA | OMA | DMO | CO_DWO | BDMSAO | EDMO |
|---|---|---|---|---|---|---|---|---|---|---|---|---|---|---|---|---|
| CEC2017-F1 | Ave | 4.9773E+03 | 1.2061E+07 | 2.5760E+10 | 1.5880E+04 | 5.7952E+05 | 6.6096E+04 | 1.6214E+10 | **1.8032E+03** | 6.2715E+03 | 4.7570E+03 | 6.8004E+08 | 4.1100E+07 | 3.9813E+07 | 7.5711E+08 | 2.7061E+03 |
| | Std | 5.2900E+03 | 1.0189E+07 | 7.9145E+09 | 2.1534E+04 | 1.4545E+06 | 8.9199E+04 | 2.2620E+09 | **9.9311E+02** | 7.1681E+03 | 5.5017E+03 | 4.0622E+08 | 2.3027E+07 | 1.9291E+07 | 6.5857E+08 | 2.7177E+03 |
| CEC2017-F3 | Ave | 6.9329E+04 | 3.0699E+04 | 3.9439E+04 | 4.7465E+05 | 6.9202E+04 | 3.2991E+04 | 9.5758E+04 | 1.3386E+04 | 2.6754E+04 | 2.1477E+04 | 5.6826E+04 | 1.4505E+05 | 1.3736E+05 | 1.5183E+05 | **3.7956E+02** |
| | Std | 1.0328E+04 | 9.9609E+03 | 8.2725E+03 | 1.9953E+05 | 5.5969E+04 | 2.3869E+04 | 7.7077E+03 | 4.8632E+03 | 6.6337E+03 | 1.4257E+04 | 1.3367E+04 | 1.8335E+04 | 2.3153E+04 | 3.0209E+04 | **1.1606E+02** |
| CEC2017-F4 | Ave | 4.9547E+02 | 5.4749E+02 | 4.1703E+03 | **4.2330E+02** | 5.0519E+02 | 5.0651E+02 | 4.0536E+03 | 4.9904E+02 | 5.0441E+02 | 5.0403E+02 | 6.5174E+02 | 5.5955E+02 | 5.6067E+02 | 7.9225E+02 | 4.8887E+02 |
| | Std | 2.7825E+01 | 4.0137E+01 | 1.5952E+03 | **1.1566E+00** | 2.5469E+01 | 2.3981E+01 | 1.1505E+03 | 1.6260E+01 | 3.5161E+01 | 3.9393E+01 | 9.7726E+01 | 2.4906E+01 | 2.2456E+01 | 1.6643E+02 | 3.0011E+01 |
| CEC2017-F5 | Ave | 5.6331E+02 | 6.0482E+02 | 7.4351E+02 | 6.0488E+02 | 5.8105E+02 | 5.8249E+02 | 7.3985E+02 | 7.3806E+02 | 6.5426E+02 | 6.4243E+02 | 6.7068E+02 | 7.3325E+02 | 7.2781E+02 | 7.5576E+02 | **5.4968E+02** |
| | Std | 1.5822E+01 | 2.9534E+01 | 3.8858E+01 | 6.7293E+01 | 1.6504E+01 | 2.3332E+01 | 2.4267E+01 | 2.3924E+01 | 3.6471E+01 | 3.9840E+01 | 3.4126E+01 | 1.4167E+01 | 1.5037E+01 | 4.8589E+01 | **1.3574E+01** |
| CEC2017-F6 | Ave | 6.0155E+02 | 6.1119E+02 | 6.5301E+02 | **6.0002E+02** | 6.0245E+02 | 6.0456E+02 | 6.6310E+02 | 6.5220E+02 | 6.2181E+02 | 6.3282E+02 | 6.2094E+02 | 6.0434E+02 | 6.0442E+02 | 6.4375E+02 | 6.0017E+02 |
| | Std | 1.6720E+00 | 5.4910E+00 | 9.9584E+00 | **7.1437E-03** | 2.2661E+00 | 1.8977E+00 | 3.8590E+00 | 4.6707E+00 | 1.0507E+01 | 9.1054E+00 | 7.3395E+00 | 7.2247E-01 | 7.8737E-01 | 1.4831E+01 | 2.5791E-01 |
| CEC2017-F7 | Ave | 8.1374E+02 | 9.0413E+02 | 1.1429E+03 | 9.1148E+02 | 8.7560E+02 | 8.3901E+02 | 1.0867E+03 | 8.6809E+02 | 9.7285E+02 | 9.9616E+02 | 1.0987E+03 | 9.7974E+02 | 9.8397E+02 | 1.1997E+03 | **7.7549E+02** |
| | Std | 2.0466E+01 | 5.4762E+01 | 7.6430E+01 | 1.5655E+01 | 3.8187E+01 | 3.0499E+01 | 4.9805E+01 | 3.7747E+01 | 6.1069E+01 | 5.6014E+01 | 7.5849E+01 | **1.5137E+01** | 1.7258E+01 | 1.0379E+02 | 1.5837E+01 |
| CEC2017-F8 | Ave | 8.7066E+02 | 8.8748E+02 | 9.9135E+02 | 9.0306E+02 | 8.8470E+02 | 8.7786E+02 | 9.6119E+02 | 9.5608E+02 | 9.4605E+02 | 9.1303E+02 | 9.5023E+02 | 1.0335E+03 | 1.0342E+03 | 1.0575E+03 | **8.4461E+02** |
| | Std | 2.1285E+01 | 2.3726E+01 | 2.3419E+01 | 7.2291E+01 | 1.9759E+01 | 1.4144E+01 | 1.9476E+01 | 2.1607E+01 | 2.8770E+01 | 2.6441E+01 | 3.2236E+01 | **1.1796E+01** | 1.2974E+01 | 3.7341E+01 | 1.4974E+01 |
| CEC2017-F9 | Ave | 1.0182E+03 | 1.9464E+03 | 5.7188E+03 | **9.0000E+02** | 1.5569E+03 | 1.3619E+03 | 4.7505E+03 | 3.4013E+03 | 3.8498E+03 | 3.1655E+03 | 2.8231E+03 | 1.9896E+03 | 1.9932E+03 | 5.2921E+03 | 9.2145E+02 |
| | Std | 1.1700E+02 | 6.9811E+02 | 1.2809E+03 | **1.6315E-02** | 3.1710E+02 | 5.6539E+02 | 3.4925E+02 | 5.2312E+02 | 9.0367E+02 | 8.6439E+02 | 1.1382E+03 | 4.3529E+02 | 3.7993E+02 | 1.5594E+03 | 1.5840E+01 |
| CEC2017-F10 | Ave | 8.2579E+03 | 4.8163E+03 | 6.0713E+03 | 9.0573E+03 | 4.5924E+03 | 5.0901E+03 | 5.1518E+03 | 5.0401E+03 | 4.5691E+03 | 5.0752E+03 | 8.2636E+03 | 8.4863E+03 | 8.5556E+03 | 5.8027E+03 | **4.4007E+03** |
| | Std | 3.9004E+02 | 6.1577E+02 | 5.5154E+02 | 3.7289E+02 | 4.2408E+02 | 7.5883E+02 | 6.2413E+02 | 6.4325E+02 | 5.8589E+02 | 5.7737E+02 | 7.4634E+02 | 3.1079E+02 | **2.8326E+02** | 6.6176E+02 | 5.2873E+02 |
| CEC2017-F11 | Ave | 1.2044E+03 | 1.2739E+03 | 1.6202E+03 | 1.1587E+04 | 1.3137E+03 | 1.3077E+03 | 7.9029E+03 | 1.1957E+03 | 1.2501E+03 | 1.3110E+03 | 1.2985E+03 | 1.6596E+03 | 1.6788E+03 | 1.9185E+03 | **1.1504E+03** |
| | Std | 3.9217E+01 | 5.8626E+01 | 1.7770E+02 | 7.3377E+03 | 2.2943E+02 | 7.0835E+01 | 1.0295E+03 | **2.4305E+01** | 5.1563E+01 | 6.0643E+01 | 5.9170E+01 | 1.1238E+02 | 1.0620E+02 | 5.7745E+02 | 2.4430E+01 |
| CEC2017-F12 | Ave | 3.2166E+05 | 1.9548E+06 | 3.4581E+08 | 5.3832E+06 | 5.9266E+05 | 4.7847E+05 | 2.7038E+09 | 4.2594E+05 | 2.7768E+06 | 3.4328E+05 | 4.9346E+06 | 2.2868E+07 | 2.4968E+07 | 2.3580E+07 | **2.0167E+05** |
| | Std | 3.4309E+05 | 1.6346E+06 | 3.8751E+08 | 5.0812E+06 | 6.9619E+05 | 3.1473E+05 | 8.8162E+08 | 1.9344E+05 | 1.9430E+06 | 5.5769E+05 | 4.0828E+06 | 7.1867E+06 | 8.3229E+06 | 2.7956E+07 | **1.7629E+05** |
| CEC2017-F13 | Ave | 1.2121E+04 | 1.7821E+04 | 3.1379E+04 | 5.5657E+06 | 2.5695E+04 | 1.8145E+04 | 4.9396E+08 | 1.4937E+04 | 2.8234E+04 | 2.0000E+04 | 2.6817E+04 | 2.6187E+04 | 2.4551E+04 | 2.1434E+05 | **5.3439E+03** |
| | Std | 1.5728E+04 | 1.9810E+04 | 2.9313E+04 | 4.1629E+06 | 1.9912E+04 | 1.6550E+04 | 5.2781E+08 | 5.7107E+03 | 2.5958E+04 | 2.1140E+04 | 1.5575E+04 | 1.6861E+04 | 1.7154E+04 | 5.9674E+05 | **3.7790E+03** |
| CEC2017-F14 | Ave | 4.4476E+03 | 4.8488E+03 | 1.5406E+03 | 5.1280E+05 | 1.6325E+03 | 1.6565E+03 | 1.4983E+06 | 5.1157E+03 | 1.9690E+05 | 2.5992E+04 | 1.7419E+04 | 1.6950E+05 | 1.4501E+05 | 5.1632E+05 | **1.5283E+03** |
| | Std | 3.6849E+03 | 7.0899E+03 | **3.6094E+01** | 3.3323E+05 | 7.6265E+01 | 1.7967E+02 | 4.8341E+05 | 3.9265E+03 | 2.0363E+05 | 2.4010E+04 | 1.9134E+04 | 1.0062E+05 | 8.4561E+04 | 6.9759E+05 | 1.9603E+02 |
| CEC2017-F15 | Ave | 7.7305E+03 | 8.4095E+03 | 3.5431E+03 | 6.0564E+06 | 4.0811E+03 | 3.0007E+03 | 1.5997E+04 | 2.7738E+03 | 1.2447E+04 | 7.6691E+03 | 1.1432E+04 | 1.1722E+04 | 1.3982E+04 | 2.3052E+04 | **1.9188E+03** |
| | Std | 6.1021E+03 | 1.0887E+04 | 1.5492E+03 | 4.2147E+06 | 4.6517E+03 | 1.1879E+03 | 3.3772E+03 | 1.2503E+03 | 1.2108E+04 | 9.2707E+03 | 7.5331E+03 | 9.4436E+03 | 9.1132E+03 | 1.9916E+04 | **6.3501E+02** |
| CEC2017-F16 | Ave | 2.7331E+03 | 2.6179E+03 | 2.7743E+03 | 2.8348E+03 | 2.6814E+03 | **2.4044E+03** | 4.2934E+03 | 3.0252E+03 | 2.6266E+03 | 2.7124E+03 | 3.2693E+03 | 3.3416E+03 | 3.4211E+03 | 2.9306E+03 | 2.4610E+03 |
| | Std | 5.6222E+02 | 3.0381E+02 | 2.0491E+02 | 2.7882E+02 | 2.3033E+02 | **1.6899E+02** | 5.4634E+02 | 3.0580E+02 | 3.7465E+02 | 3.3195E+02 | 2.1251E+02 | 1.8908E+02 | 1.9981E+02 | 3.1397E+02 | 2.8835E+02 |

| ID | Metric | QANA | SMO | IAO | CMA_ES | LSHADE | LSHADE_cnEpSin | GSA | IPO | MSMA | GWCA | OMA | DMO | CO_DWO | BDMSAO | EDMO |
|---|---|---|---|---|---|---|---|---|---|---|---|---|---|---|---|---|
| CEC2017-F17 | Ave | 1.9277E+03 | 2.1451E+03 | 2.0254E+03 | 2.2918E+03 | 2.0374E+03 | **1.9211E+03** | 2.9944E+03 | 2.7043E+03 | 2.4573E+03 | 2.3452E+03 | 2.1762E+03 | 2.4148E+03 | 2.4422E+03 | 2.3800E+03 | 1.9319E+03 |
|  | Std | 1.2647E+02 | 2.1214E+02 | 1.0710E+02 | 2.1639E+02 | 1.5297E+02 | **9.9927E+01** | 1.8098E+02 | 2.2148E+02 | 2.0698E+02 | 2.5202E+02 | 1.1275E+02 | 1.2316E+02 | 1.2396E+02 | 2.2110E+02 | 1.5927E+02 |
| CEC2017-F18 | Ave | 7.0032E+05 | 1.4099E+05 | **3.2486E+03** | 6.2654E+06 | 1.0222E+05 | 4.4571E+04 | 3.6878E+06 | 1.1109E+05 | 2.5939E+06 | 2.9630E+05 | 2.9797E+05 | 5.4779E+06 | 5.9524E+06 | 3.0671E+06 | 2.9982E+04 |
|  | Std | 4.7196E+05 | 1.6209E+05 | **2.0116E+03** | 3.3642E+06 | 1.7450E+05 | 3.4003E+04 | 3.0677E+06 | 3.1301E+04 | 2.6106E+06 | 3.8612E+05 | 2.4634E+05 | 2.3579E+06 | 3.1619E+06 | 3.9732E+06 | 1.6993E+04 |
| CEC2017-F19 | Ave | 1.2250E+04 | 1.3281E+04 | 2.5710E+03 | 2.0333E+06 | 7.4919E+03 | **2.2966E+03** | 1.1533E+06 | 5.2005E+03 | 7.6851E+03 | 1.0555E+04 | 1.0705E+04 | 1.3951E+04 | 2.0494E+04 | 3.3307E+04 | 3.5920E+03 |
|  | Std | 1.2759E+04 | 1.3593E+04 | 1.7901E+03 | 1.4635E+06 | 1.4003E+04 | **7.5047E+02** | 1.2589E+06 | 1.3921E+03 | 8.3110E+03 | 1.3544E+04 | 1.0340E+04 | 1.1150E+04 | 1.3331E+04 | 4.9226E+04 | 3.8845E+03 |
| CEC2017-F20 | Ave | 2.3349E+03 | 2.4251E+03 | 2.3563E+03 | 2.6560E+03 | 2.4615E+03 | 2.3208E+03 | 3.1845E+03 | 2.8187E+03 | 2.5078E+03 | 2.7492E+03 | 2.6065E+03 | 2.7439E+03 | 2.7986E+03 | 2.5977E+03 | **2.2704E+03** |
|  | Std | 1.8769E+02 | 1.7191E+02 | **6.7652E+01** | 1.5583E+02 | 1.1498E+02 | 1.0426E+02 | 2.2188E+02 | 2.3511E+02 | 1.5592E+02 | 2.0911E+02 | 1.2761E+02 | 1.2295E+02 | 1.3610E+02 | 1.8272E+02 | 1.3859E+02 |
| CEC2017-F21 | Ave | 2.3602E+03 | 2.3903E+03 | 2.4965E+03 | 2.4016E+03 | 2.3784E+03 | 2.3775E+03 | 2.6630E+03 | 2.5585E+03 | 2.4446E+03 | 2.4197E+03 | 2.4579E+03 | 2.5268E+03 | 2.5245E+03 | 2.5459E+03 | **2.3443E+03** |
|  | Std | 1.5828E+01 | 2.0261E+01 | 3.0821E+01 | 6.9032E+01 | 1.9948E+01 | 1.9970E+01 | 4.5090E+01 | 3.6060E+01 | 3.2538E+01 | 2.8334E+01 | 3.5987E+01 | **9.8495E+00** | 1.5971E+01 | 2.3759E+01 | 1.4567E+01 |
| CEC2017-F22 | Ave | 2.3008E+03 | 2.8749E+03 | 5.2591E+03 | 1.0306E+04 | 3.8317E+03 | 3.3352E+03 | 7.4966E+03 | 6.4933E+03 | 4.5914E+03 | 4.5662E+03 | 2.5065E+03 | 4.8419E+03 | 5.2877E+03 | 6.6161E+03 | **2.3004E+03** |
|  | Std | 1.4946E+00 | 1.4120E+01 | 9.6022E+02 | 3.2102E+02 | 1.9307E+03 | 1.8323E+03 | 5.2321E+02 | 1.6196E+03 | 1.9622E+03 | 2.3237E+03 | 1.6067E+02 | 1.9994E+03 | 2.1364E+03 | 1.8094E+03 | **1.1586E+00** |
| CEC2017-F23 | Ave | **2.7070E+03** | 2.7571E+03 | 2.9391E+03 | 2.7192E+03 | 2.7427E+03 | 2.7463E+03 | 3.9668E+03 | 3.3751E+03 | 2.7750E+03 | 2.8417E+03 | 2.8419E+03 | 2.8735E+03 | 2.8744E+03 | 2.9082E+03 | 2.7071E+03 |
|  | Std | 1.3134E+01 | 2.5756E+01 | 7.8006E+01 | 5.9763E+01 | 2.4102E+01 | 2.1862E+01 | 2.1104E+02 | 1.5717E+02 | 3.6041E+01 | 6.2258E+01 | 3.4471E+01 | **1.1079E+01** | 1.4154E+01 | 3.8032E+01 | 2.2218E+01 |
| CEC2017-F24 | Ave | 2.8901E+03 | 2.9171E+03 | 3.1054E+03 | 2.8951E+03 | 2.8972E+03 | 2.9099E+03 | 4.0367E+03 | 3.3154E+03 | 2.9478E+03 | 3.0007E+03 | 3.0523E+03 | 3.0333E+03 | 3.0334E+03 | 3.0787E+03 | **2.8742E+03** |
|  | Std | 4.0787E+01 | 3.4742E+01 | 5.7805E+01 | 6.2035E+01 | 1.8970E+01 | 2.5229E+01 | 1.5361E+02 | 7.2383E+01 | 3.4130E+01 | 6.6006E+01 | 4.6845E+01 | 1.2801E+01 | **1.0505E+01** | 4.1081E+01 | 1.8317E+01 |
| CEC2017-F25 | Ave | 2.8959E+03 | 2.9446E+03 | 3.5837E+03 | **2.8784E+03** | 2.9058E+03 | 2.8961E+03 | 3.2852E+03 | 2.9225E+03 | 2.9043E+03 | 2.9067E+03 | 3.0417E+03 | 2.9221E+03 | 2.9215E+03 | 3.1486E+03 | 2.8883E+03 |
|  | Std | 1.7780E+01 | 2.9917E+01 | 1.9260E+02 | **1.1348E-01** | 2.0914E+01 | 1.3083E+01 | 7.7441E+01 | 1.6626E+01 | 1.8075E+01 | 2.4169E+01 | 5.1256E+01 | 1.0247E+01 | 9.7001E+00 | 1.3606E+02 | 4.4337E+00 |
| CEC2017-F26 | Ave | 4.3146E+03 | 4.8051E+03 | 6.9507E+03 | **4.1333E+03** | 4.7542E+03 | 4.6111E+03 | 8.9710E+03 | 6.9899E+03 | 5.0368E+03 | 5.5277E+03 | 5.8090E+03 | 5.9375E+03 | 5.9518E+03 | 6.3169E+03 | 4.3047E+03 |
|  | Std | 3.3053E+02 | 7.8058E+02 | 1.1683E+03 | 4.1028E+02 | 5.9526E+02 | 3.1248E+02 | 3.6854E+02 | 1.6711E+03 | 3.0788E+02 | 1.1934E+03 | 3.0743E+02 | 1.2527E+02 | **1.0098E+02** | 8.9109E+02 | 2.0308E+02 |
| CEC2017-F27 | Ave | 3.2201E+03 | 3.2505E+03 | 3.2889E+03 | **3.2000E+03** | 3.2367E+03 | 3.2364E+03 | 5.3257E+03 | 4.1769E+03 | 3.2363E+03 | 3.2957E+03 | 3.2973E+03 | 3.2364E+03 | 3.2340E+03 | 3.2936E+03 | 3.2190E+03 |
|  | Std | 1.4757E+01 | 2.0578E+01 | 3.9963E+01 | **7.8457E-05** | 1.4045E+01 | 1.8768E+01 | 4.4141E+02 | 2.7970E+02 | 2.0308E+01 | 4.7435E+01 | 3.1933E+01 | 4.8317E+00 | 5.3225E+00 | 2.8765E+01 | 1.4308E+01 |
| CEC2017-F28 | Ave | 3.2285E+03 | 3.3156E+03 | 4.3833E+03 | 3.3000E+03 | 3.2531E+03 | 3.2673E+03 | 4.5308E+03 | 3.2317E+03 | 3.2647E+03 | 3.2418E+03 | 3.4850E+03 | 3.3341E+03 | 3.3294E+03 | 3.5886E+03 | **3.1889E+03** |
|  | Std | 2.0113E+01 | 5.1670E+01 | 4.0919E+02 | **7.3702E-05** | 2.6900E+01 | 4.9386E+01 | 2.3006E+02 | 2.1293E+01 | 3.5332E+01 | 2.5264E+01 | 9.3986E+01 | 2.5727E+01 | 2.3305E+01 | 2.0155E+02 | 3.3745E+01 |
| CEC2017-F29 | Ave | 3.6738E+03 | 3.8834E+03 | 4.1379E+03 | 3.9728E+03 | 3.7630E+03 | 3.6242E+03 | 6.6817E+03 | 4.4495E+03 | 3.9916E+03 | 4.3106E+03 | 4.1207E+03 | 4.4519E+03 | 4.4579E+03 | 4.4566E+03 | **3.5749E+03** |
|  | Std | 1.6818E+02 | 2.0604E+02 | 2.4400E+02 | 2.2667E+02 | 1.1573E+02 | **1.0456E+02** | 5.4074E+02 | 2.4320E+02 | 2.1876E+02 | 2.8884E+02 | 1.6564E+02 | 1.5366E+02 | 1.5612E+02 | 3.4737E+02 | 1.3474E+02 |
| CEC2017-F30 | Ave | 1.3842E+04 | 3.2653E+04 | 1.0146E+05 | 1.8306E+06 | 1.9730E+04 | 3.0403E+04 | 3.3345E+07 | 1.1011E+04 | 1.6922E+04 | 9.2988E+03 | 1.0129E+05 | 5.7450E+05 | 5.1515E+05 | 2.4727E+05 | 9.0496E+03 |
|  | Std | 5.5972E+03 | 3.2256E+04 | 9.0840E+04 | 9.7289E+05 | 8.3700E+03 | 1.5999E+04 | 3.7550E+07 | 3.3761E+03 | 6.0639E+03 | 2.4049E+03 | 1.2909E+05 | 4.9650E+05 | 2.7647E+05 | 1.8026E+05 | 3.2752E+03 |
| Friedman average |  | 4.53 | 6.50 | 9.83 | 8.12 | 5.33 | 4.74 | 13.67 | 8.22 | 7.06 | 7.28 | 9.44 | 10.59 | 10.70 | 11.67 | **2.33** |
| Overall Rank |  | 2 | 5 | 11 | 8 | 4 | 3 | 15 | 9 | 6 | 7 | 10 | 12 | 13 | 14 | **1** |

Table 6. Test results for CEC2017 (d=100).

| ID | Metric | QANA | SMO | IAO | CMA_ES | LSHADE | LSHADE_cnEpSin | GSA | IPO | MSMA | GWCA | OMA | DMO | CO_DWO | BDMSAO | EDMO |
|---|---|---|---|---|---|---|---|---|---|---|---|---|---|---|---|---|
| CEC2017-F1 | Ave | 6.2865E+02 | 6.5243E+02 | 6.9057E+02 | **6.0448E+02** | 6.3733E+02 | 6.3900E+02 | 6.7341E+02 | 6.6062E+02 | 6.6213E+02 | 6.5954E+02 | 6.7692E+02 | 6.6620E+02 | 6.6601E+02 | 6.8195E+02 | 6.2027E+02 |
|  | Std | 4.5206E+00 | 7.4039E+00 | 4.7958E+00 | **4.6537E-01** | 8.1686E+00 | 4.5826E+00 | 2.4800E+00 | 5.3938E+00 | 3.8641E+00 | 4.5979E+00 | 5.3049E+00 | 3.4568E+00 | 3.2149E+00 | 8.9802E+00 | 4.4783E+00 |
| CEC2017-F3 | Ave | 1.7545E+03 | 3.0900E+03 | 3.5517E+03 | 1.7286E+03 | 2.9076E+03 | 2.5105E+03 | 3.1656E+03 | 1.7777E+03 | 2.8152E+03 | 3.1764E+03 | 3.5665E+03 | 4.5572E+03 | 4.5137E+03 | 4.3038E+03 | **1.4428E+03** |
|  | Std | 1.5348E+02 | 3.3005E+02 | 1.4570E+02 | **1.9592E+01** | 3.5821E+02 | 2.6472E+02 | 1.4976E+02 | 8.8414E+01 | 2.1177E+02 | 2.7367E+02 | 3.5551E+02 | 2.1429E+02 | 2.0491E+02 | 4.6831E+02 | 1.1843E+02 |
| CEC2017-F4 | Ave | 1.3060E+03 | 1.6843E+03 | 2.2747E+03 | 1.7048E+03 | 1.6294E+03 | 1.5816E+03 | 1.9912E+03 | 1.7658E+03 | 1.6783E+03 | 1.6189E+03 | 1.9035E+03 | 2.1016E+03 | 2.1022E+03 | 2.1826E+03 | **1.1681E+03** |
|  | Std | 9.9527E+01 | 1.0178E+02 | 8.6847E+01 | **2.4065E+01** | 1.0501E+02 | 1.2710E+02 | 6.3382E+01 | 4.0621E+01 | 8.3693E+01 | 1.0188E+02 | 1.2187E+02 | 3.3845E+01 | 2.7742E+01 | 1.5490E+02 | 4.3412E+01 |
| CEC2017-F5 | Ave | 5.1601E+04 | 3.1751E+04 | 5.7478E+04 | **1.0934E+03** | 3.6437E+04 | 3.5550E+04 | 2.9432E+04 | 2.1500E+04 | 2.7195E+04 | 2.4250E+04 | 6.6853E+04 | 7.3753E+04 | 7.6058E+04 | 6.4277E+04 | 7.9221E+03 |
|  | Std | 1.4104E+04 | 5.6286E+03 | 2.8045E+03 | **5.8786E+01** | 7.0115E+03 | 7.9938E+03 | 2.2784E+03 | 9.7239E+02 | 2.7854E+03 | 3.3259E+03 | 7.0983E+03 | 7.1578E+03 | 6.8371E+03 | 1.2144E+04 | 1.8590E+03 |
| CEC2017-F6 | Ave | 3.2191E+04 | 2.0961E+04 | 2.6717E+04 | 3.4056E+04 | 2.1574E+04 | 2.4968E+04 | 2.1784E+04 | 1.6119E+04 | 1.5727E+04 | 1.7306E+04 | 3.1888E+04 | 3.2488E+04 | 3.2538E+04 | 2.5340E+04 | **1.4655E+04** |
|  | Std | 7.8624E+02 | 1.6205E+03 | 9.3533E+02 | 6.3988E+02 | 9.2738E+02 | 3.9007E+03 | 1.4537E+03 | 1.4274E+03 | 1.3377E+03 | 1.7088E+03 | 8.4389E+02 | **6.2138E+02** | 6.5487E+02 | 1.9613E+03 | 1.3009E+03 |
| CEC2017-F7 | Ave | 1.4474E+05 | 5.6084E+04 | 1.0415E+05 | 1.0400E+06 | 5.2521E+04 | 4.2110E+04 | 1.9860E+05 | 1.7193E+04 | 3.1206E+04 | 2.6992E+04 | 8.6964E+04 | 2.6888E+05 | 2.5630E+05 | 2.3875E+05 | **6.0042E+03** |
|  | Std | 2.2992E+04 | 1.4351E+04 | 2.1194E+04 | 3.1388E+05 | 2.4341E+04 | 3.1126E+04 | 3.1413E+04 | 5.2188E+03 | 8.1049E+03 | 1.9301E+04 | 1.2709E+04 | 3.6207E+04 | 2.4998E+04 | 4.8284E+04 | **2.1593E+03** |
| CEC2017-F8 | Ave | 6.6413E+07 | 2.7689E+09 | 1.2860E+11 | 2.0653E+09 | 7.3195E+08 | 6.4204E+08 | 1.5451E+11 | 7.0077E+07 | 1.3099E+08 | 7.1633E+08 | 1.3038E+10 | 1.1419E+10 | 1.1501E+10 | 2.7805E+09 | **1.8948E+07** |
|  | Std | 3.7753E+07 | 1.2314E+09 | 2.1531E+10 | 8.0746E+08 | 5.6754E+08 | 3.1165E+08 | 9.9975E+09 | 3.3556E+07 | 6.6911E+07 | 1.4127E+09 | 3.9987E+09 | 1.6341E+09 | 1.4490E+09 | 9.5013E+08 | **8.8047E+06** |
| CEC2017-F9 | Ave | 1.5842E+04 | 1.4005E+07 | 2.3525E+10 | 3.6231E+08 | 6.3664E+05 | 8.4560E+04 | 3.2615E+10 | 1.8230E+04 | 1.2179E+05 | 3.6228E+06 | 3.2944E+08 | 4.2862E+06 | 3.9993E+06 | 9.3448E+05 | **5.6653E+03** |
|  | Std | 6.2474E+03 | 3.3126E+07 | 6.5595E+09 | 1.2395E+08 | 9.7493E+05 | 2.3430E+04 | 2.8371E+09 | **3.5366E+03** | 2.4332E+05 | 1.9549E+07 | 2.4558E+08 | 2.1937E+06 | 1.6051E+06 | 3.4586E+06 | 4.8686E+03 |
| CEC2017-F10 | Ave | 5.5897E+06 | 2.5769E+06 | 2.0669E+06 | 7.6104E+07 | 1.5669E+06 | 5.7953E+05 | 2.7301E+07 | 9.3376E+05 | 2.5858E+06 | 1.3840E+06 | 2.8089E+06 | 5.6470E+07 | 5.1836E+07 | 1.6596E+07 | **4.7390E+05** |
|  | Std | 3.5187E+06 | 1.4189E+06 | 1.6106E+06 | 2.4589E+07 | 1.3017E+06 | 4.8295E+05 | 1.0601E+07 | 2.0522E+05 | 1.1889E+06 | 7.8672E+05 | 1.2962E+06 | 1.2038E+07 | 1.0934E+07 | 9.2915E+06 | **1.7951E+05** |
| CEC2017-F11 | Ave | 6.4321E+03 | 3.9062E+04 | 6.0198E+09 | 3.4910E+08 | 2.4768E+04 | 4.8846E+04 | 1.5666E+10 | 9.2871E+03 | 3.1586E+04 | 7.0300E+03 | 6.6139E+06 | 2.8921E+04 | 3.1679E+04 | 8.0924E+05 | **3.3100E+03** |
|  | Std | 5.6239E+03 | 4.2187E+04 | 3.9583E+09 | 1.1915E+08 | 1.3558E+04 | 2.0471E+04 | 2.3109E+09 | 2.1095E+03 | 9.1081E+04 | 3.6382E+03 | 6.3504E+06 | 1.8924E+04 | 4.0545E+04 | 3.9563E+06 | **1.8580E+03** |
| CEC2017-F12 | Ave | 6.9193E+03 | 6.5752E+03 | 1.2991E+04 | 1.1018E+04 | 7.3892E+03 | 6.6085E+03 | 1.8197E+04 | 6.6295E+03 | 6.3164E+03 | 5.9477E+03 | 8.7063E+03 | 1.1415E+04 | 1.1476E+04 | 1.0177E+04 | **5.3017E+03** |
|  | Std | 2.2150E+03 | 4.4977E+02 | 1.9588E+03 | 6.4876E+02 | 6.7726E+02 | 9.2077E+02 | 1.4901E+03 | 5.8496E+02 | 7.2328E+02 | 5.9871E+02 | 1.2591E+03 | **3.2766E+02** | 3.3450E+02 | 9.6835E+02 | 6.2037E+02 |
| CEC2017-F13 | Ave | 5.5029E+03 | 5.6363E+03 | 3.0085E+04 | 1.5433E+04 | 5.9388E+03 | 5.2581E+03 | 6.3617E+06 | 5.0759E+03 | 5.8243E+03 | 6.0632E+03 | 5.9438E+03 | 8.0811E+03 | 8.2042E+03 | 7.0362E+03 | **4.4269E+03** |
|  | Std | 1.1203E+03 | 5.1974E+02 | 3.7574E+04 | 1.6885E+03 | 6.0218E+02 | 4.6214E+02 | 4.9102E+06 | 4.5464E+02 | 6.9826E+02 | 6.3473E+02 | 5.6395E+02 | 3.3891E+02 | **2.4841E+02** | 7.2080E+02 | 4.6357E+02 |
| CEC2017-F14 | Ave | 8.7154E+06 | 5.1398E+06 | 2.1467E+06 | 1.2038E+08 | 2.1881E+06 | 1.1429E+06 | 5.5543E+07 | 1.1340E+06 | 8.0346E+06 | 3.1902E+06 | 4.5320E+06 | 1.0009E+08 | 1.0325E+08 | 2.7967E+07 | **7.6630E+05** |
|  | Std | 4.6946E+06 | 3.6818E+06 | 1.3616E+06 | 4.1775E+07 | 1.7875E+06 | 8.5903E+05 | 3.7780E+07 | **2.9688E+05** | 2.8459E+06 | 1.8655E+06 | 1.9811E+06 | 2.2852E+07 | 2.5097E+07 | 1.8008E+07 | 3.8129E+05 |
| CEC2017-F15 | Ave | 6.7673E+03 | 3.6496E+05 | 6.3718E+09 | 3.5989E+08 | 5.2763E+04 | 5.2024E+05 | 1.5491E+10 | 5.4002E+03 | 1.5895E+04 | 7.3412E+03 | 1.1929E+07 | 4.8863E+05 | 3.4472E+05 | 1.5772E+06 | **3.5654E+03** |
|  | Std | 7.6292E+03 | 5.1478E+05 | 3.7317E+09 | 9.2248E+07 | 7.1695E+04 | 6.1274E+05 | 2.1031E+09 | 2.1161E+03 | 2.5925E+04 | 5.9865E+03 | 8.9041E+06 | 7.6775E+05 | 2.8312E+05 | 2.1764E+06 | **1.8334E+03** |
| CEC2017-F16 | Ave | 7.6068E+03 | 5.5944E+03 | 5.6791E+03 | 8.0786E+03 | 6.3797E+03 | 6.0898E+03 | 6.5774E+03 | 5.6225E+03 | 5.3506E+03 | 5.4143E+03 | 7.4050E+03 | 7.9210E+03 | 7.9565E+03 | 5.8990E+03 | **4.6479E+03** |
|  | Std | 2.5637E+02 | 5.4904E+02 | 3.2896E+02 | 3.6817E+02 | 6.3661E+02 | 5.1489E+02 | 5.4870E+02 | 5.7443E+02 | 5.5963E+02 | 5.9400E+02 | 2.2894E+02 | 2.6616E+02 | **2.0405E+02** | 6.4031E+02 | 4.9439E+02 |

| ID | Metric | QANA | SMO | IAO | CMA_ES | LSHADE | LSHADE_cnEpSin | GSA | IPO | MSMA | GWCA | OMA | DMO | CO_DWO | BDMSAO | EDMO |
|---|---|---|---|---|---|---|---|---|---|---|---|---|---|---|---|---|
| CEC2017-F17 | Ave | 2.7874E+03 | 3.2619E+03 | 3.9992E+03 | 3.2376E+03 | 3.1633E+03 | 3.1411E+03 | 5.3655E+03 | 3.7124E+03 | 3.1716E+03 | 3.2725E+03 | 3.3095E+03 | 3.6198E+03 | 3.6253E+03 | 3.7021E+03 | **2.6824E+03** |
| | Std | 9.1230E+01 | 1.0848E+02 | 1.3125E+02 | **2.9812E+01** | 1.1419E+02 | 1.1906E+02 | 2.1574E+02 | 1.4058E+02 | 1.1371E+02 | 1.1525E+02 | 1.1012E+02 | 3.2473E+01 | 3.6580E+01 | 1.4590E+02 | 4.6375E+01 |
| CEC2017-F18 | Ave | 3.4679E+04 | 2.3434E+04 | 2.9906E+04 | 3.5635E+04 | 2.4636E+04 | 2.7290E+04 | 2.5149E+04 | 2.0612E+04 | 1.8622E+04 | 2.0345E+04 | 3.4584E+04 | 3.4770E+04 | 3.4825E+04 | 2.7707E+04 | **1.7364E+04** |
| | Std | 6.9691E+02 | 1.3430E+03 | 7.3327E+02 | 9.4398E+02 | 1.5237E+03 | 3.5635E+03 | 1.1357E+03 | 1.3162E+03 | 1.3152E+03 | 1.1874E+03 | 6.3299E+02 | 4.5423E+02 | **3.4382E+02** | 2.5951E+03 | 1.1774E+03 |
| CEC2017-F19 | Ave | 3.3096E+03 | 3.8475E+03 | 4.6897E+03 | 3.7653E+03 | 3.6572E+03 | 3.8053E+03 | 8.2869E+03 | 5.7999E+03 | 3.4139E+03 | 4.1808E+03 | 3.9464E+03 | 4.0203E+03 | 4.0202E+03 | 3.8943E+03 | **3.2465E+03** |
| | Std | 7.9492E+01 | 1.2734E+02 | 1.6141E+02 | 3.6517E+01 | 9.5337E+01 | 1.5705E+02 | 6.0633E+02 | 4.2697E+02 | 7.7620E+01 | 2.7917E+02 | 1.0607E+02 | **2.6569E+01** | 2.9956E+01 | 1.4176E+02 | 6.1068E+01 |
| CEC2017-F20 | Ave | **3.8124E+03** | 4.7090E+03 | 6.2765E+03 | 4.1848E+03 | 4.3708E+03 | 4.5513E+03 | 1.3303E+04 | 4.8398E+03 | 4.1892E+03 | 5.1624E+03 | 5.5100E+03 | 4.5059E+03 | 4.5037E+03 | 4.6684E+03 | 3.8233E+03 |
| | Std | 8.5432E+01 | 2.2343E+02 | 3.9086E+02 | 3.4373E+01 | 1.9002E+02 | 2.2821E+02 | 5.9856E+02 | 2.1463E+02 | 1.4120E+02 | 4.6863E+02 | 2.9147E+02 | 3.6390E+01 | **3.3839E+01** | 1.9540E+02 | 1.0828E+02 |
| CEC2017-F21 | Ave | 3.7145E+03 | 6.8229E+03 | 2.1161E+04 | **3.3581E+03** | 5.1202E+03 | 4.3127E+03 | 2.0258E+04 | 3.8060E+03 | 3.9092E+03 | 4.6522E+03 | 1.0166E+04 | 1.8538E+04 | 1.8601E+04 | 1.0327E+04 | 3.4329E+03 |
| | Std | 8.5864E+01 | 9.1213E+02 | 1.3336E+03 | **4.1139E+01** | 5.6823E+02 | 2.5546E+02 | 1.4173E+03 | 6.8902E+01 | 1.1592E+02 | 4.4723E+02 | 1.2712E+03 | 1.4819E+03 | 1.7243E+03 | 3.1114E+03 | 5.2142E+01 |
| CEC2017-F22 | Ave | 1.2190E+04 | 2.0194E+04 | 4.4067E+04 | 1.5182E+04 | 1.6857E+04 | 1.8242E+04 | 4.6658E+04 | 2.4942E+04 | 1.6792E+04 | 2.3393E+04 | 3.0128E+04 | 1.8687E+04 | 1.8693E+04 | 2.1014E+04 | **1.1995E+04** |
| | Std | 2.1593E+03 | 3.0495E+03 | 2.9201E+03 | 3.6224E+02 | 1.5773E+03 | 1.8831E+03 | 2.4637E+03 | 2.5867E+03 | 2.7722E+03 | 3.3812E+03 | 3.2413E+03 | 4.0661E+02 | **3.5355E+02** | 1.3802E+03 | 9.1642E+02 |
| CEC2017-F23 | Ave | 3.6795E+03 | 4.2058E+03 | 5.5360E+03 | **3.2000E+03** | 3.9933E+03 | 4.0912E+03 | 1.6264E+04 | 1.0100E+04 | 3.7395E+03 | 4.1716E+03 | 5.0135E+03 | 4.7610E+03 | 4.7852E+03 | 4.3241E+03 | 3.6972E+03 |
| | Std | 1.1252E+02 | 2.2599E+02 | 5.4952E+02 | **6.6923E-05** | 2.2440E+02 | 1.8738E+02 | 1.3584E+03 | 1.2622E+03 | 1.2875E+02 | 2.6451E+02 | 2.8829E+02 | 1.3214E+02 | 1.2413E+02 | 4.1127E+02 | 9.9972E+01 |
| CEC2017-F24 | Ave | 3.8336E+03 | 1.0160E+04 | 2.4970E+04 | **3.3000E+03** | 6.8315E+03 | 5.5854E+03 | 2.8107E+04 | 4.2828E+03 | 3.7524E+03 | 5.9787E+03 | 1.3403E+04 | 1.8392E+04 | 1.8459E+04 | 1.4300E+04 | 3.5943E+03 |
| | Std | 1.6786E+02 | 1.1360E+03 | 2.9481E+03 | **6.9126E-05** | 1.5165E+03 | 1.1453E+03 | 1.5670E+03 | 1.5911E+02 | 7.6672E+01 | 1.1492E+03 | 1.7437E+03 | 7.7770E+02 | 8.7195E+02 | 2.6046E+03 | 4.4534E+01 |
| CEC2017-F25 | Ave | 6.9563E+03 | 8.7709E+03 | 2.5921E+04 | 1.6957E+04 | 8.4874E+03 | 8.4558E+03 | 3.5571E+05 | 8.2957E+03 | 7.5931E+03 | 8.5473E+03 | 1.1257E+04 | 1.1454E+04 | 1.1283E+04 | 1.2029E+04 | **6.2230E+03** |
| | Std | 5.5817E+02 | 7.5843E+02 | 1.1598E+04 | 1.4785E+03 | 1.0628E+03 | 6.6813E+02 | 1.3734E+05 | 4.6564E+02 | 5.5865E+02 | 8.4582E+02 | 1.3416E+03 | 4.0143E+02 | **3.7255E+02** | 1.5567E+03 | 5.0647E+02 |
| CEC2017-F26 | Ave | 1.6199E+05 | 2.7625E+07 | 1.0907E+10 | 5.8272E+08 | 4.1928E+06 | 1.7537E+07 | 3.0182E+10 | 5.4613E+05 | 7.2174E+05 | 5.0963E+05 | 6.1707E+08 | 3.7651E+07 | 3.7429E+07 | 4.2936E+07 | **3.0682E+04** |
| | Std | 9.1841E+04 | 2.3679E+07 | 5.0944E+09 | 1.2995E+08 | 2.6931E+06 | 1.2190E+07 | 2.6447E+09 | 3.0271E+05 | 4.2231E+05 | 1.2849E+06 | 7.6682E+08 | 1.2706E+07 | 9.7146E+06 | 5.1115E+07 | **1.6786E+04** |
| CEC2017-F27 | Ave | 6.2865E+02 | 6.5243E+02 | 6.9057E+02 | **6.0448E+02** | 6.3733E+02 | 6.3900E+02 | 6.7341E+02 | 6.6062E+02 | 6.6213E+02 | 6.5954E+02 | 6.7692E+02 | 6.6620E+02 | 6.6601E+02 | 6.8195E+02 | 6.2027E+02 |
| | Std | 4.5206E+00 | 7.4039E+00 | 4.7958E+00 | **4.6537E-01** | 8.1686E+00 | 4.5826E+00 | 2.4800E+00 | 5.3938E+00 | 3.8641E+00 | 4.5979E+00 | 5.3049E+00 | 3.4568E+00 | 3.2149E+00 | 8.9802E+00 | 4.4783E+00 |
| CEC2017-F28 | Ave | 1.7545E+03 | 3.0900E+03 | 3.5517E+03 | 1.7286E+03 | 2.9076E+03 | 2.5105E+03 | 3.1656E+03 | 1.7777E+03 | 2.8152E+03 | 3.1764E+03 | 3.5665E+03 | 4.5572E+03 | 4.5137E+03 | 4.3038E+03 | **1.4428E+03** |
| | Std | 1.5348E+02 | 3.3005E+02 | 1.4570E+02 | **1.9592E+01** | 3.5821E+02 | 2.6472E+02 | 1.4976E+02 | 8.8414E+01 | 2.1177E+02 | 2.7367E+02 | 3.5551E+02 | 2.1429E+02 | 2.0491E+02 | 4.6831E+02 | 1.1843E+02 |
| CEC2017-F29 | Ave | 1.3060E+03 | 1.6843E+03 | 2.2747E+03 | 1.7048E+03 | 1.6294E+03 | 1.5816E+03 | 1.9912E+03 | 1.7658E+03 | 1.6783E+03 | 1.6189E+03 | 1.9035E+03 | 2.1016E+03 | 2.1022E+03 | 2.1826E+03 | **1.1681E+03** |
| | Std | 9.9527E+01 | 1.0178E+02 | 8.6847E+01 | **2.4065E+01** | 1.0501E+02 | 1.2710E+02 | 6.3382E+01 | 4.0621E+01 | 8.3693E+01 | 1.0188E+02 | 1.2187E+02 | 3.3845E+01 | 2.7742E+01 | 1.5490E+02 | 4.3412E+01 |
| CEC2017-F30 | Ave | 5.1601E+04 | 3.1751E+04 | 5.7478E+04 | **1.0934E+03** | 3.6437E+04 | 3.5550E+04 | 2.9432E+04 | 2.1500E+04 | 2.7195E+04 | 2.4250E+04 | 6.6853E+04 | 7.3753E+04 | 7.6058E+04 | 6.4277E+04 | 7.9221E+03 |
| | Std | 1.4104E+04 | 5.6286E+03 | 2.8045E+03 | **5.8786E+01** | 7.0115E+03 | 7.9938E+03 | 2.2784E+03 | 9.7239E+02 | 2.7854E+03 | 3.3259E+03 | 7.0983E+03 | 7.1578E+03 | 6.8371E+03 | 1.2144E+04 | 1.8590E+03 |
| Friedman average | | 4.95 | 7.45 | 11.97 | 8.54 | 6.21 | 5.98 | 12.67 | 5.64 | 4.81 | 6.15 | 10.52 | 11.59 | 11.56 | 10.37 | 1.59 |
| Overall Rank | | 3 | 8 | 14 | 9 | 7 | 5 | 15 | 4 | 2 | 6 | 11 | 13 | 12 | 10 | **1** |

Table 7. Test results for CEC2020 (d=10).

| ID | Metric | QANA | SMO | IAO | CMA_ES | LSHADE | LSHADE_cnEpSin | GSA | IPO | MSMA | GWCA | OMA | DMO | CO_DWO | BDMSAO | EDMO |
|---|---|---|---|---|---|---|---|---|---|---|---|---|---|---|---|---|
| CEC2020-F1 | Ave | 2.6972E+03 | 2.7266E+03 | 2.5745E+03 | 2.7441E+03 | 2.7229E+03 | 2.7297E+03 | 2.6511E+03 | 2.7425E+03 | 2.7322E+03 | 2.7527E+03 | 2.6825E+03 | 2.7233E+03 | 2.6837E+03 | 2.7719E+03 | **2.5324E+03** |
| | Std | 9.3472E+01 | 6.2032E+01 | 1.1266E+02 | 1.2773E+01 | 6.0723E+01 | 4.3520E+01 | 1.7873E+02 | 1.6344E+02 | 7.9264E+01 | **1.2103E+01** | 8.8901E+01 | 5.1844E+01 | 5.8230E+01 | 2.8953E+01 | 8.8070E+01 |
| CEC2020-F2 | Ave | 2.9266E+03 | 2.9266E+03 | 2.9149E+03 | 2.9396E+03 | 2.9288E+03 | 2.9246E+03 | 2.9406E+03 | 2.9373E+03 | 2.9295E+03 | 2.9304E+03 | 2.9222E+03 | 2.9238E+03 | 2.9331E+03 | 2.9631E+03 | **2.9058E+03** |
| | Std | 2.3294E+01 | 2.2981E+01 | 2.2313E+01 | **1.4108E+01** | 2.3986E+01 | 2.3123E+01 | 1.4666E+01 | 1.7304E+01 | 2.3865E+01 | 2.4582E+01 | 2.2540E+01 | 1.6559E+01 | 1.4894E+01 | 2.1453E+01 | 1.7522E+01 |
| CEC2020-F3 | Ave | 7.2083E+02 | 7.2356E+02 | 7.2694E+02 | 7.3876E+02 | 7.1736E+02 | **7.1640E+02** | 7.2756E+02 | 7.3063E+02 | 7.3496E+02 | 7.3710E+02 | 7.4750E+02 | 7.4115E+02 | 7.3997E+02 | 7.8631E+02 | 7.1916E+02 |
| | Std | 4.9163E+00 | 8.9451E+00 | 7.8120E+00 | 3.9115E+00 | 4.3123E+00 | **3.2304E+00** | 8.5062E+00 | 7.2239E+00 | 1.0122E+01 | 1.0918E+01 | 1.0936E+01 | 5.3396E+00 | 4.7677E+00 | 1.7003E+01 | 5.4651E+00 |
| CEC2020-F4 | Ave | 1.9009E+03 | 1.9012E+03 | 1.9014E+03 | 1.9021E+03 | **1.9008E+03** | 1.9009E+03 | 1.9044E+03 | 1.9009E+03 | 1.9022E+03 | 1.9020E+03 | 1.9031E+03 | 1.9027E+03 | 1.9026E+03 | 1.9068E+03 | 1.9008E+03 |
| | Std | 3.4490E-01 | 5.8696E-01 | 5.3221E-01 | 4.4821E-01 | **2.3655E-01** | 2.5678E-01 | 1.9955E+00 | 2.5145E-01 | 7.8588E-01 | 1.0571E+00 | 1.0954E+00 | 3.5896E-01 | 4.1135E-01 | 2.6453E+00 | 3.3139E-01 |
| CEC2020-F5 | Ave | 2.0472E+03 | 1.9854E+03 | 1.7557E+03 | 1.6983E+05 | 1.9959E+03 | 2.0159E+03 | 3.9362E+05 | 3.2362E+03 | 1.0680E+05 | 5.3916E+03 | 4.6868E+03 | 1.2684E+04 | 8.1640E+03 | 1.1034E+05 | **1.7240E+03** |
| | Std | 6.9279E+02 | 2.2427E+02 | 4.3443E+01 | 1.4376E+05 | 2.2632E+02 | 1.7558E+02 | 2.3907E+05 | 7.5610E+02 | 1.4310E+05 | 2.8502E+03 | 1.6908E+03 | 7.0740E+03 | 5.3515E+03 | 1.6837E+05 | **2.2852E+01** |
| CEC2020-F6 | Ave | 1.6003E+03 | 1.6005E+03 | **1.6003E+03** | 1.6025E+03 | 1.6005E+03 | 1.6005E+03 | 1.6689E+03 | 1.6007E+03 | 1.6005E+03 | 1.6008E+03 | 1.6006E+03 | 1.6005E+03 | 1.6004E+03 | 1.6010E+03 | 1.6004E+03 |
| | Std | 2.8761E-01 | 3.6516E-01 | **2.8606E-01** | 1.9408E+00 | 3.7185E-01 | 3.5118E-01 | 4.6484E+01 | 5.2884E-01 | 3.3814E-01 | 3.9403E-01 | 3.3040E-01 | 3.6052E-01 | 2.9958E-01 | 6.3235E-01 | 3.2169E-01 |
| CEC2020-F7 | Ave | 2.2575E+03 | 2.2468E+03 | **2.1070E+03** | 6.9798E+04 | 2.2099E+03 | 2.2235E+03 | 5.9079E+05 | 6.5013E+03 | 6.7939E+03 | 7.0707E+03 | 4.1730E+03 | 5.7642E+03 | 4.8278E+03 | 1.6053E+04 | 2.1224E+03 |
| | Std | 1.7611E+02 | 3.7619E+02 | **2.3005E+01** | 7.0858E+04 | 9.3649E+01 | 1.2322E+02 | 4.4462E+05 | 1.7700E+03 | 4.7145E+03 | 5.6919E+03 | 2.0016E+03 | 1.9706E+03 | 1.0649E+03 | 1.1510E+04 | 2.6848E+01 |
| CEC2020-F8 | Ave | 2.2963E+03 | 2.2949E+03 | 2.2976E+03 | 3.4652E+03 | 2.2986E+03 | 2.2980E+03 | 2.3041E+03 | 2.3027E+03 | 2.2994E+03 | 2.3038E+03 | 2.2988E+03 | 2.3017E+03 | 2.2965E+03 | 2.3296E+03 | **2.2840E+03** |
| | Std | 1.9046E+01 | 2.1984E+01 | 1.5518E+01 | 8.1659E+02 | 1.2300E+01 | 1.3689E+01 | 6.8805E+00 | **1.6197E+00** | 1.3258E+01 | 3.1790E+00 | 1.5324E+01 | 3.0840E+00 | 1.4938E+01 | 2.5314E+01 | 3.4169E+01 |
| CEC2020-F9 | Ave | 2.6846E+03 | 2.7342E+03 | **2.5625E+03** | 2.7440E+03 | 2.7388E+03 | 2.7285E+03 | 2.6333E+03 | 2.7287E+03 | 2.7422E+03 | 2.7492E+03 | 2.7087E+03 | 2.7324E+03 | 2.6837E+03 | 2.7676E+03 | 2.5634E+03 |
| | Std | 1.0367E+02 | 4.4559E+01 | 1.0542E+02 | 1.4277E+01 | **4.2350E+00** | 4.3241E+01 | 1.8431E+02 | 1.6660E+02 | 6.7271E+01 | 4.8329E+01 | 6.1569E+01 | 5.0612E+01 | 6.6680E+01 | 4.7920E+01 | 1.0699E+02 |
| CEC2020-F10 | Ave | 2.9234E+03 | 2.9238E+03 | 2.9121E+03 | 2.9364E+03 | 2.9199E+03 | 2.9181E+03 | 2.9454E+03 | 2.9327E+03 | 2.9343E+03 | 2.9387E+03 | 2.9134E+03 | 2.9238E+03 | 2.9316E+03 | 2.9590E+03 | **2.8990E+03** |
| | Std | 2.3409E+01 | 2.3915E+01 | 2.1150E+01 | 1.7453E+01 | 2.3557E+01 | 2.2972E+01 | **2.3491E+00** | 2.0459E+01 | 3.0903E+01 | 2.6097E+01 | 2.1226E+01 | 2.0291E+01 | 1.5791E+01 | 1.4310E+01 | 5.9734E+01 |
| Friedman average | | 5.47 | 6.57 | 4.50 | 10.61 | 5.30 | 4.93 | 11.04 | 8.95 | 9.90 | 9.89 | 8.72 | 9.33 | 8.59 | 13.29 | **2.91** |
| Overall Rank | | 5 | 6 | 2 | 13 | 4 | 3 | 14 | 9 | 12 | 11 | 8 | 10 | 7 | 15 | **1** |

The experimental results in Table 5, Table 6 and Table 7 demonstrate that, compared to the latest 14 algorithms, EDMO consistently achieves the best or near-best performance across most test functions and never exhibits the worst performance on any function. The comparative performance of the algorithms was rigorously assessed using the non-parametric Friedman average rank test. EDMO consistently achieved the highest ranking (1st) in both the CEC2017 and CEC2020 benchmark suites, indicating its statistically superior performance with the lowest average rank among all compared algorithms. Furthermore, EDMO demonstrated significantly bolded values in means and standard deviations across multiple test functions, highlighting its exceptional solution accuracy and robustness. These statistically validated results unequivocally establish EDMO's outstanding optimization potential in the analyzed test scenarios, outperforming all competing algorithms in both convergence precision and stability.

To further analyze the convergence speed and iterative process of the aforementioned algorithms, 15 different types of test functions were selected for comparison. As shown in Figure 5, algorithms like GSA, CMA_ES, DMO, and BDMSAO tend to get trapped in local optima after a certain number of iterations, preventing them from continuing the search for the global optimum. Specifically, GSA's fixed gravitational parameters restrict its adaptability, causing inefficient exploration in dynamic landscapes, while CMA-ES's covariance matrix adaptation, though effective in unimodal scenarios, struggles with high-dimensional multimodal problems. Similarly, DMO and BDMSAO, due to their rigid parameter control and lack of adaptive search strategies, fail to balance exploration and exploitation effectively. WOA's sinusoidal search behavior may lead to stagnation in certain optimization landscapes, while LSHADE's adaptive mutation strategies, though effective in some cases, may not efficiently handle high-dimensional multimodal problems. In contrast, EDMO's convergence curve rapidly decreases during the early and middle stages of iteration, demonstrating superior speed and accuracy compared to other algorithms. This advantage is attributed to strategies such as OLOBL and adaptive update mechanisms, which guide the algorithm in efficiently exploring different regions of the solution space, identifying candidate solutions, and conducting the search process more effectively, leading to faster convergence towards the optimal solution.

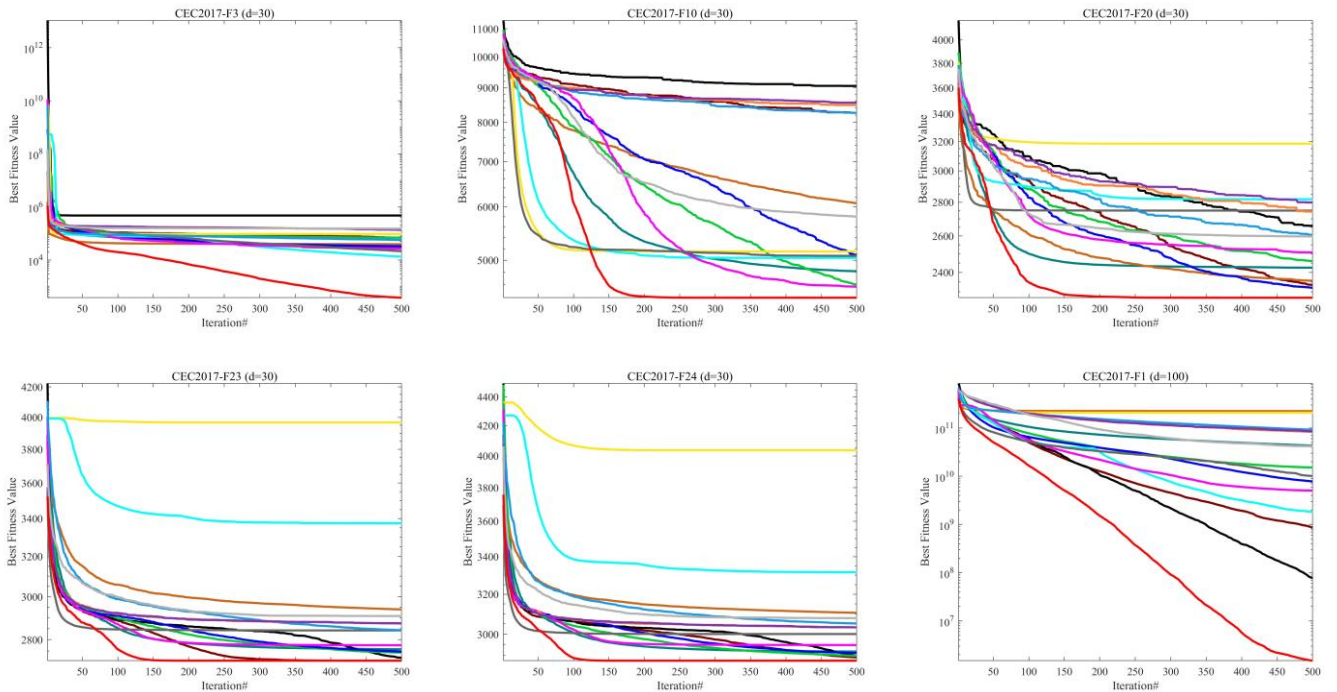

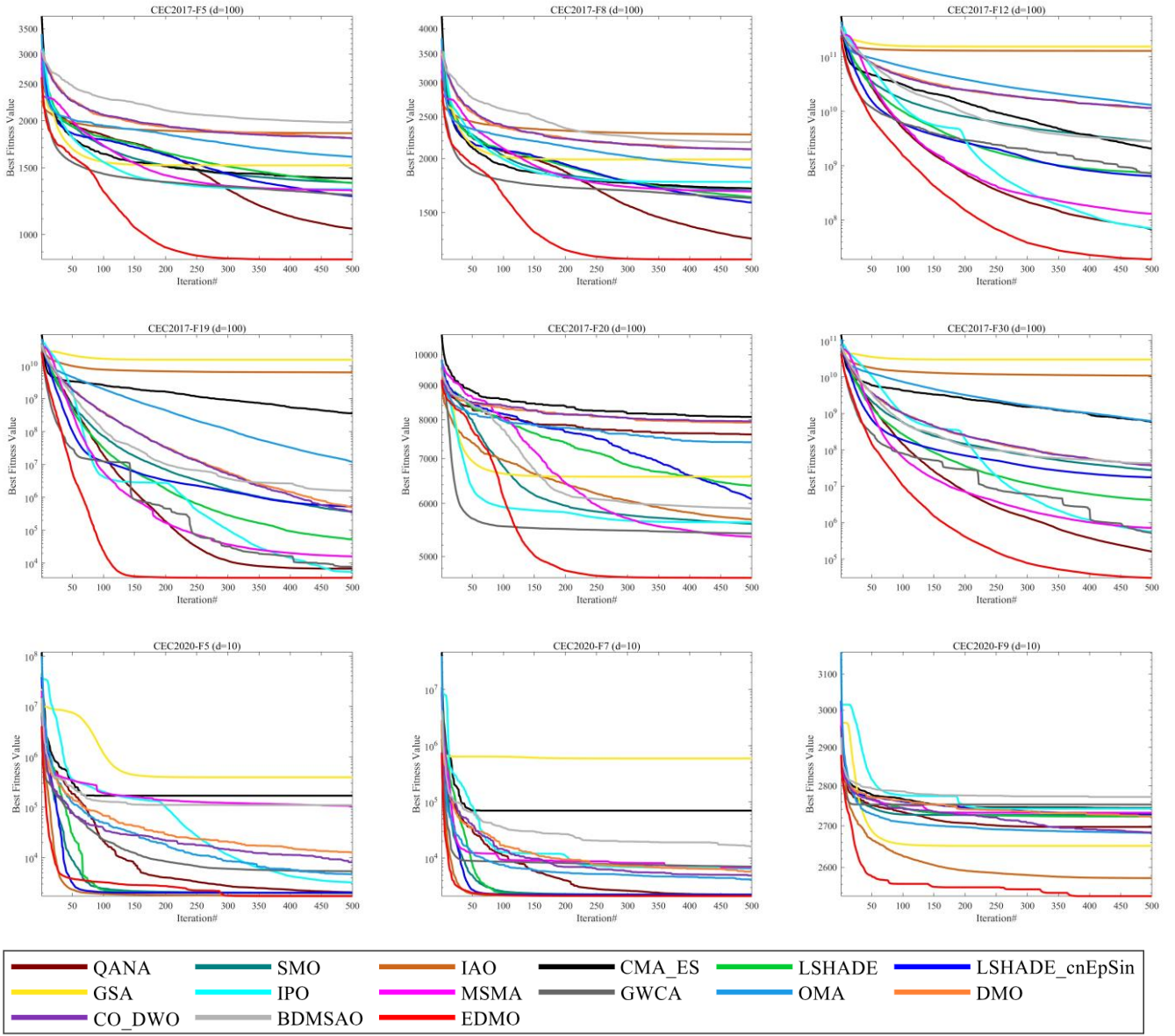

Figure 5. Convergence curve comparison.

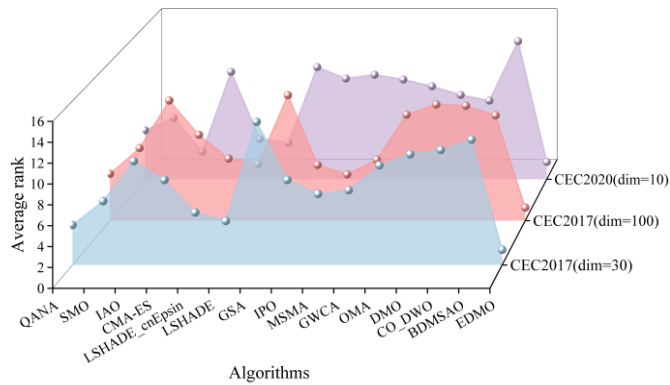

Figure 6. EDMO and comparison algorithm average ranking.

In addition, to provide a more intuitive view of each algorithm's performance, the experimental results from Table 5, Table 6 and Table 7 are presented in graphical form, as shown in Figure 6. From this figure, it is clear that EDMO outperforms all competing optimizers in terms of average ranking on the CEC2017 (d = 30, 100)and CEC2020 (d = 10) test functions, with scores of 1.50, 1.33 and 1.80, respectively. In contrast, the original DMO ranked 12th, 13th and 9th with scores of 10.80, 11.73 and 8.30, respectively, while LSHADE and LSHADE_cnEpsin scored 5.13, 6.27, 4.20 and 4.30, 5.73, 3.80, respectively. These results demonstrate the effectiveness of the four strategies embedded in DMO, enabling EDMO to achieve high-quality solutions with strong stability, faster convergence speed, higher precision, and a robust ability to avoid being trapped in local optima.

# 6 EDMO algorithm practical engineering application

In this section, the practical application of the EDMO algorithm is explored using the UAV three-dimensional path planning problem and three complex design challenges, including Pressure Vessel Design (PVD) problem [63], the Tension/Compression Spring Design (T-BTD) problem [63], the Cantilever Beam Design (CBD) problem [64]. By simulating the UAV's flight path in complex mountainous models, the outstanding application potential of the EDMO algorithm in addressing the currently popular path planning issues is demonstrated. By solving three engineering design problems, EDMO has great application potential in engineering problems.

## 6.1 The engineering design problems

### 6.1.1 Pressure vessel design problem (PVD)

The Pressure Vessel Design (PVD) problem aims to find an optimal balance between cost minimization and structural integrity. Key parameters such as head thickness $(T_h)$, cylinder length $(L)$, shell thickness $(T_s)$, and inner radius $(R)$ are adjusted to optimize the design. Figure 7 provides a structural illustration of this pressure vessel, clearly depicting its main components and dimensions. Additionally, we present a mathematical formulation of this problem, offering researchers a specific and quantifiable reference model for further analysis and optimization.

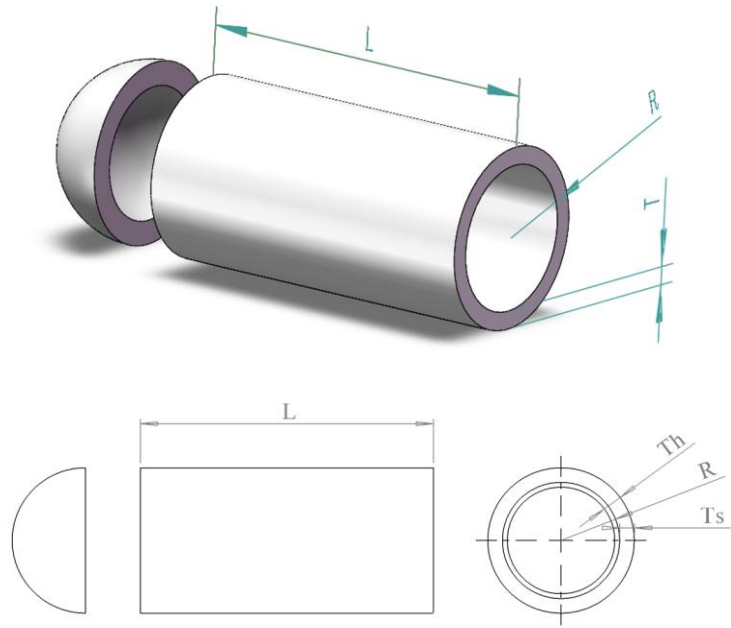

Figure 7. Schematic representation of the PVD.

Parameters range:

| | |
|---|---|
| Consider: | $\vec{x}=[x_1\ \ x_2\ \ x_3\ \ x_4]=[T_s\ \ T_h\ \ R\ \ L]$ |
| Minimize: | $f(\vec{x})=0.6224x_1x_3x_4+1.7781x_2x_3^2+3.1661x_1^2x_4\ 19.84x_1^2x_3$ |
| Subject to: | $g_1(\vec{x})=-x_1+0.0193\leq0$ |
| | $g_2(\vec{x})=-x_3+0.00954\leq0$ |
| | $g_3(\vec{x})=-\pi x_3^2-\frac{4}{3}\pi x_3^3+129600\leq0$ |
| | $g_4(\vec{x})=x_4-240\leq0$ |
| Parameters range: | $0 \leq x_1, x_2 \leq 99$, $10 \leq x_3, x_4 \leq 200$ |

Table 8 provides a comprehensive evaluation of 15 optimization algorithms applied to the Pressure Vessel Design problem, with the EDMO algorithm consistently achieving the optimal design cost. The low mean and standard deviation of EDMO's results highlight its cost efficiency as well as its solution reliability. These findings emphasize EDMO's effectiveness in solving complex engineering challenges. Furthermore, the Wilcoxon rank-sum test confirms EDMO's superior performance compared to other algorithms, reinforcing its significant advantage in tackling such optimization tasks.

### 6.1.2 Tension/compression spring design problem (T/CSD)

In the T/CSD problem, the objective is to minimize spring mass by judiciously optimizing three parameters: coil diameter (D), wire diameter (d), and number of active coils (n), subject to design constraints. Figure 8 visualizes the spring structure, and we delineates the optimization model.

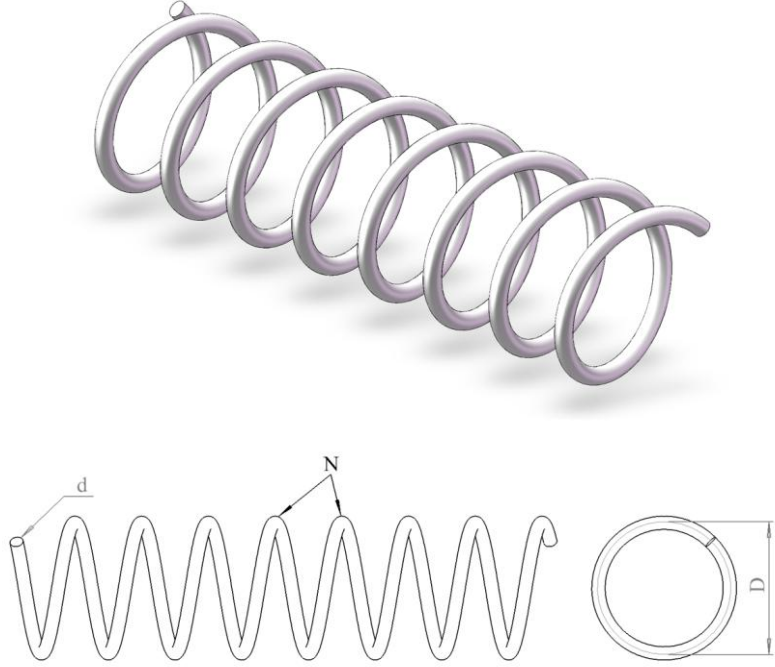

Figure 8. Schematic representation of the T/CSD.

| | |
|---|---|
| Consider: | $\vec{x} = [x_1\ x_2\ x_3] = [d\ N\ D]$ |
| Minimize: | $f(\vec{x}) = (x_3+2) + 1.7781 x_2 x_1^2$ |
| Subject to: | $g_1(\vec{x}) = 1 - \frac{x_2 x_1^2}{71785 x_1^4} \leq 0$ |
| | $g_2(\vec{x}) = 0.00954 \frac{4x_2^2 - x_1 x_2}{12566(x_2 x_1^3 - x_1^4)} + \frac{1}{5108 x_1^2} \leq 0$ |
| | $g_3(\vec{x}) = 1 - \frac{140.45 x_1}{x_3 x_2^2} \leq 0$ |
| | $g_4(\vec{x}) = \frac{x_1 + x_2}{1.5} - 1 \leq 0,$ |
| Parameters range: | $0 \leq x_1 \leq 2$, $0.25 \leq x_2 \leq 1.3$, $2 \leq x_3 \leq 15$ |

Table 9 presents a comparative analysis of EDMO against 14 other prominent algorithms, evaluating them across a range of metrics. Although EDMO shares the lowest average value with OMA and CO_DWO, it excels in terms of standard deviation, which is significantly lower than the other algorithms, demonstrating its superior performance and robustness. The results of the Wilcoxon rank-sum test further validate EDMO's significant outperformance over its competitors, reinforcing its proficiency in addressing complex optimization challenges.

## 6.1.3 Cantilever beam design problem (CBD)

The CBD problem focuses on optimizing a five-cube hollow structure, where height variations aim to minimize costs under specific constraints. Designers must judiciously adjust each cube's height to ensure structural efficiency and cost-effectiveness. Figure 9 illustrates the structure, and we mathematically articulates the optimization problem, defining the objective function and constraints to guide the cost-efficient design process.

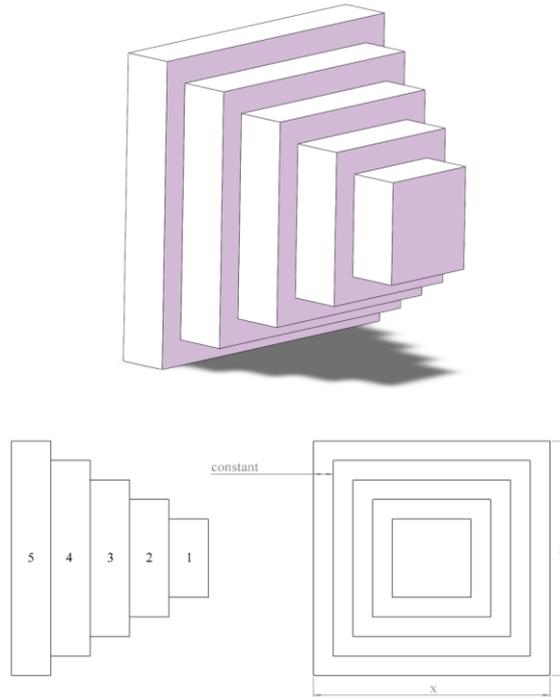

Figure 9. Schematic representation of the CBD.

| | |
|---:|:---|
| Consider: | $\vec{x}=[x_1 \quad x_2 \quad x_3 \quad x_4 \quad x_5]$ |
| Minimize: | $f(\vec{x})=0.6224(x_1+x_2+x_3+x_4+x_5)$ |
| Subject to: | $g(\vec{x})=\dfrac{60}{x_1^3}+\dfrac{27}{x_2^3}+\dfrac{19}{x_3^3}+\dfrac{7}{x_4^3}+\dfrac{1}{x_5^3}-1\leq 0$ |
| Parameter range: | $0.01\leq x_1,x_2,x_3,x_4,x_5\leq 100$ |

Table 10 highlights EDMO's dominance in optimizing the CBD problem, delivering superior results across all evaluation metrics, which underscores its efficacy and consistency. The algorithm's low mean and standard deviation further emphasize its stable solution discovery. The Wilcoxon rank-sum test corroborates EDMO's significant performance advantage over competitors, solidifying its robustness in handling such complex engineering challenges.

Table 8. Comparative results for PVD.

| Algorithm | QANA | SMO | IAO | CMA-ES | LSHADE | LSHADE_cnEpsin | GSA | IPO | MSMA | GWCA | OMA | DMO | CO_DWO | BDMSAO | EDMO |
|---|---|---|---|---|---|---|---|---|---|---|---|---|---|---|---|
| Ave | 6.0399E+03 | 6.0150E+03 | 5.9917E+03 | 2.6055E+04 | 8.9869E+03 | 1.0967E+04 | 2.2945E+05 | 6.2631E+03 | 6.2188E+03 | 6.3597E+03 | 5.9384E+03 | 6.0315E+03 | 6.0067E+03 | 9.8085E+03 | **5.8853E+03** |
| Std | 1.4434E+02 | 1.3104E+02 | 1.2167E+02 | 1.2680E+04 | 1.3009E+03 | 2.6615E+03 | 9.3496E+04 | 7.4125E+01 | 4.3768E+02 | 4.4498E+02 | 3.7823E+01 | 1.2302E+02 | 8.2666E+01 | 3.1904E+03 | **1.5669E-02** |
| Rank | 7 | 4 | 3 | 14 | 11 | 13 | 15 | 10 | 8 | 9 | 2 | 5 | 6 | 12 | **1** |
| Wilcoxion | (+) | (+) | (+) | (+) | (+) | (+) | (+) | (+) | (+) | (+) | (+) | (+) | (+) | (+) | (=) |

Table 9. Comparative results for T/CSD.

| Algorithm | QANA | SMO | IAO | CMA-ES | LSHADE | LSHADE_cnEpsin | GSA | IPO | MSMA | GWCA | OMA | DMO | CO_DWO | BDMSAO | EDMO |
|---|---|---|---|---|---|---|---|---|---|---|---|---|---|---|---|
| Ave | 1.2674E-02 | 1.2681E-02 | 1.2680E-02 | 1.9106E-02 | 1.4424E-02 | 1.5410E-02 | 1.9827E-02 | 7.6036E+12 | 1.2836E-02 | 1.3809E-02 | 1.2731E-02 | 1.2804E-02 | 1.2788E-02 | 1.5068E-02 | **1.2665E-02** |
| Std | 1.2496E-05 | 1.5948E-05 | 1.4650E-05 | 4.0571E-03 | 8.0054E-04 | 1.9957E-03 | 4.6607E-03 | 4.1647E+13 | 2.5602E-04 | 1.5140E-03 | 3.3709E-05 | 1.0014E-04 | 9.7677E-05 | 2.1740E-03 | **1.0395E-07** |
| Rank | 2 | 3 | 4 | 14 | 11 | 13 | 15 | 5 | 7 | 10 | 6 | 9 | 8 | 12 | **1** |
| Wilcoxion | (+) | (+) | (+) | (+) | (+) | (+) | (+) | (+) | (+) | (+) | (+) | (+) | (+) | (+) | (=) |

Table 10. Comparative results for CBD.

| Algorithm | QANA | SMO | IAO | CMA-ES | LSHADE | LSHADE_cnEpsin | GSA | IPO | MSMA | GWCA | OMA | DMO | CO_DWO | BDMSAO | EDMO |
|---|---|---|---|---|---|---|---|---|---|---|---|---|---|---|---|
| Ave | 6.3464E+03 | 6.3367E+03 | 6.3167E+03 | 2.4782E+04 | 8.6857E+03 | 1.0661E+04 | 1.8883E+05 | 6.6818E+03 | 6.8802E+03 | 6.9197E+03 | 6.2844E+03 | 6.3725E+03 | 6.3504E+03 | 8.2188E+03 | **6.2770E+03** |
| Std | 1.2287E+02 | 8.8005E+01 | 7.7831E+01 | 1.0889E+04 | 9.5583E+02 | 1.9848E+03 | 5.9596E+04 | 1.1193E+02 | 5.0192E+02 | 4.3543E+02 | 6.5754E+00 | 9.5569E+01 | 7.5538E+01 | 1.6991E+03 | **3.8436E-04** |
| Rank | 5 | 4 | 2 | 14 | 12 | 13 | 15 | 9 | 8 | 10 | 3 | 7 | 6 | 11 | 1 |
| Wilcoxion | (+) | (+) | (+) | (+) | (+) | (+) | (+) | (+) | (+) | (+) | (+) | (+) | (+) | (+) | (=) |

## 6.2 UAV path planning model
### 6.2.1 Flight distance cost

The flight distance cost primarily accounts for the fuel consumption of the UAV during its flight. When it is assumed that the UAV reaches a specified speed during the mission and maintains constant velocity flight, the fuel consumption becomes directly proportional to the total flight distance of the UAV [16]. The formula for calculating the flight distance cost is as follows:

$$f_{range} = \frac{\varepsilon}{Q_r} \sum_{i=1}^{n} L_i \quad (22)$$

where, $\varepsilon$ represents the fuel consumption per unit flight distance, $Q_r$ represents the total amount of fuel carried by the UAV, and $L_i$ indicates the length of the $i^{th}$ flight segment.

### 6.2.2 Flight altitude cost

During UAV flight, the likelihood of the aircraft being affected by low temperatures is directly related to the flight altitude, whereas the risk of terrain threats is inversely proportional to the altitude. To ensure UAV safety, we have established a maximum flight altitude $h_{max}$ and a minimum flight altitude $h_{min}$. Assuming the current flight altitude is $h_i$, the cost function for flight altitude can be expressed as:

$$f_{range} = \begin{cases} \frac{h_i - h_{min}}{h_{max} - h_{min}}, & h_{min} < h_i < h_{max} \\ \infty, & others \end{cases} \quad (23)$$

$$f_{altitude} = \frac{1}{n} \sum_{i=1}^{n} f_{h_i} \quad (24)$$

### 6.2.3 Flight risk cost

When planning the UAV's flight path, it is crucial to account for environmental risks such as terrain, extreme weather, military operations, and other hazardous areas. Suppose that, in the $i^{th}$ f section of the planned flight path, the UAV approaches a flight risk point $k$. If this segment $L_i$ is divided into $m$ sub-segments, the flight risk cost function for segment $L_i$ due to risk point $k$ can be expressed as:

$$f_{n_i,k} = \frac{1}{m} \left( P_k(d_{k,1}) + P_k(d_{k,2}) + \cdots + P_k(d_{k,m}) \right) \quad (25)$$

where, $n_i$ represents the $i^{th}$ segment $L_i$ in the trajectory planning; $P_k$ represents the destruction probability of the UAV by the kth threat point; $d_{k,m}$ represents the distance from the $k^{th}$ threat point to the $m^{th}$ sub-segment in segment $L_i$. The total threat cost incurred by the UAV's flight path planning can be expressed as:

$$f_{risk} = \frac{1}{n}\frac{1}{k} \sum_{i=1}^{n} \sum_{i}^{k} f_{n_i,k} \quad (26)$$

### 6.2.4 Performance measurement function of Multi-UAV collaborative path planning

Considering the comprehensive factors of flight distance cost, flight altitude cost, and flight risk cost in UAV trajectory planning, the influence of each cost on the trajectory planning will vary. Assuming their weight coefficients are $\omega_1$, $\omega_2$, and $\omega_3$ respectively, the objective function for the collaborative trajectory planning of multiple UAV can be expressed as follows:

$$f = \omega_1 \times f_{range} + \omega_2 \times f_{altitude} + \omega_3 \times f_{risk} \quad (27)$$

## 6.3 Simulation and analysis of UAV 3D path planning
### 6.3.1 Algorithm application and experimental simulation

To illustrate the practical application of the proposed algorithm, we conducted simulation tests on the 3D UAV path using the MATLAB platform. The dimensions of the simulation scenario are 100 km × 150 km × 30 km, with the starting point at (10, 90) and the target point at (130, 10). Table 11 provides the specific two-dimensional coordinate parameters of the risk areas. In the simulation of threat zones, these areas are highlighted in pink, as shown in Figure 10.

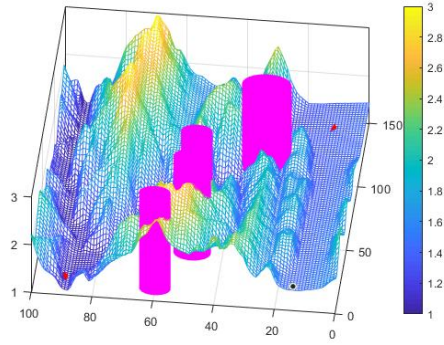

Figure 10. Three-dimensional threat topographic map.

Table 11. 2-D coordinate parameters of risk area.

| Risky Area | Center Point | Risk Radius |
| --- | --- | --- |
| 1 | (10,60) | 5 |
| 2 | (40,50) | 6 |
| 3 | (60,50) | 5 |
| 4 | (100,30) | 8 |

The DMO algorithm was employed to address the aforementioned problem for 3D path planning as a reference, with the parameter settings as follows: population size $N = 30$ and maximum iteration number $T = 200$. The parameter settings for the EDMO algorithm are consistent with those outlined in Section 4.2. The simulation outcomes for the two-dimensional path plan are illustrated in Figure 11, while the results for the three-dimensional path plan are also displayed in Figure 12. A comparative analysis of the convergence curves for the six algorithms is presented in 13.

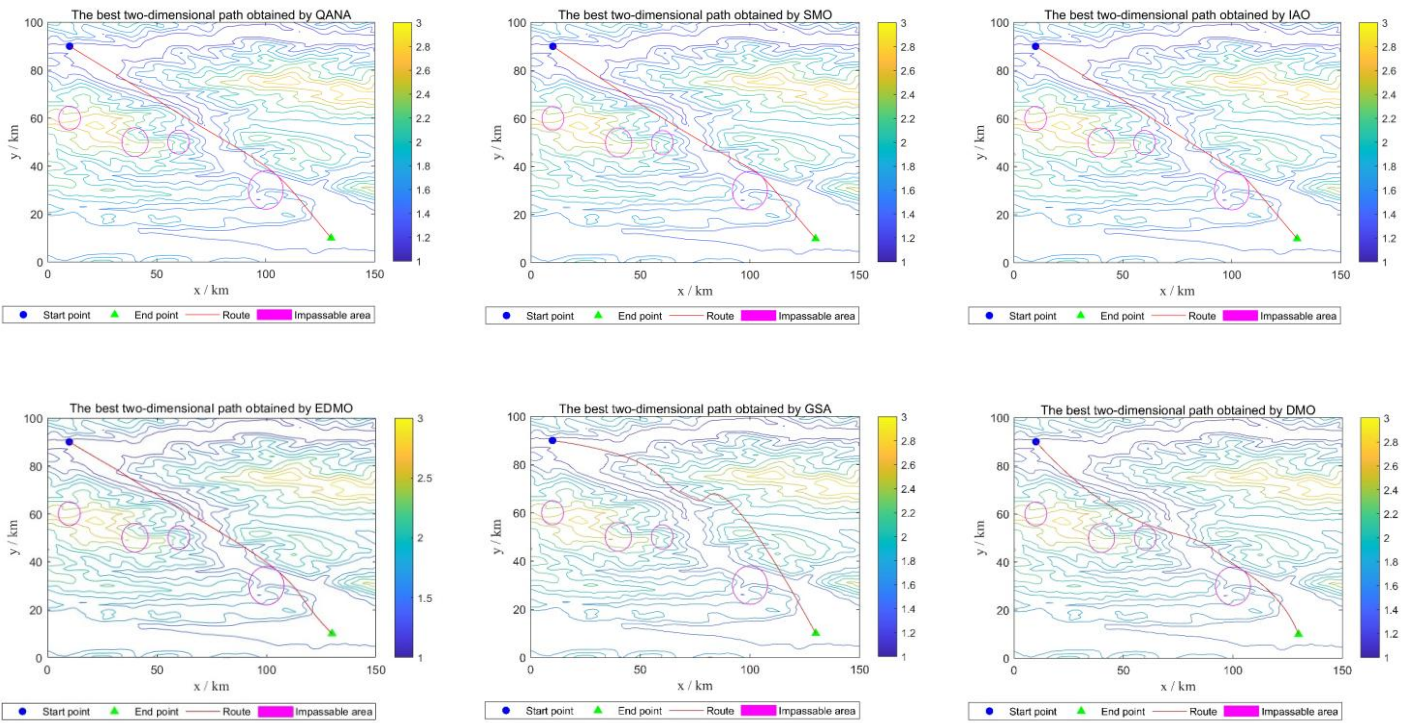

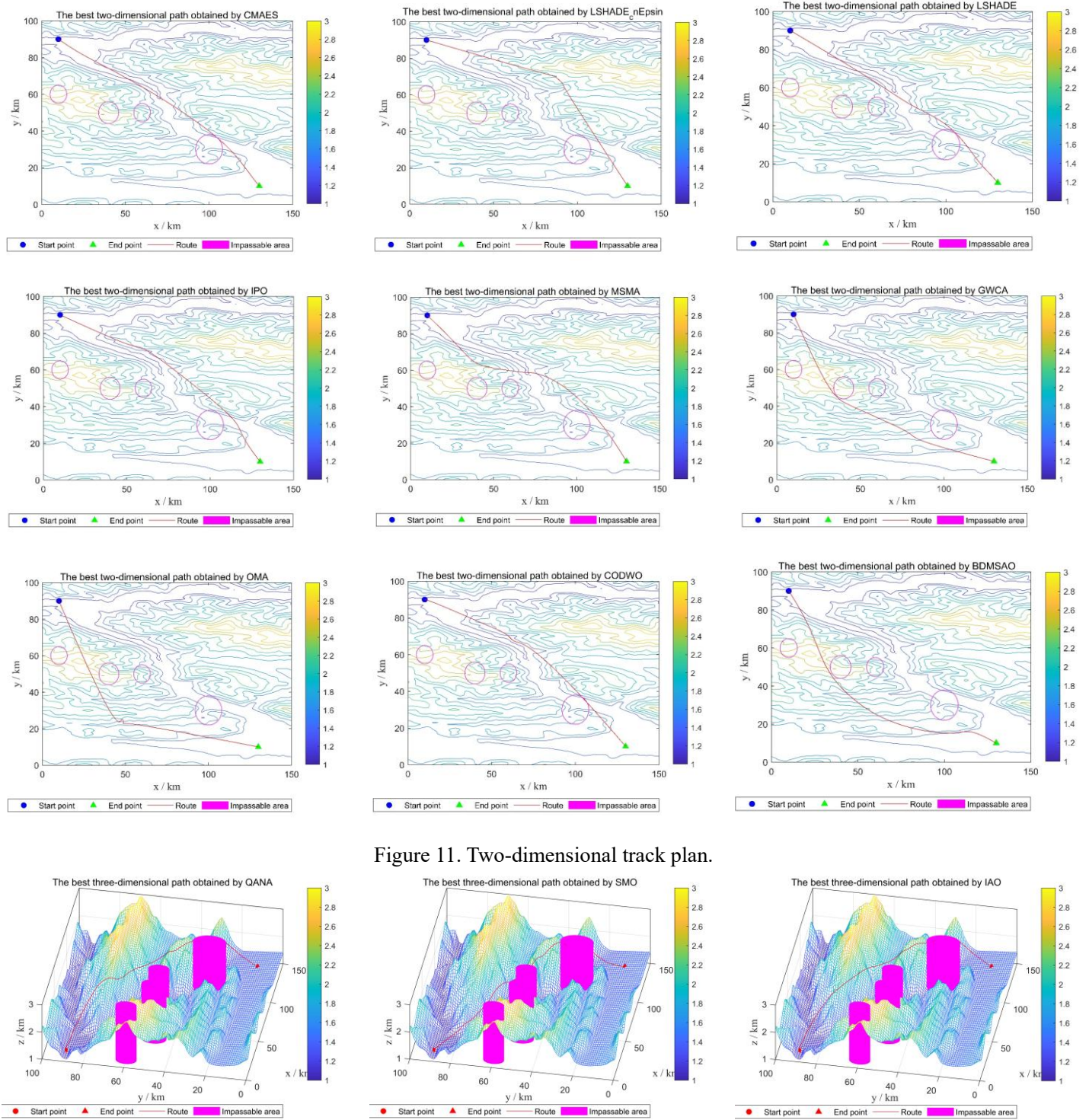

Figure 11. Two-dimensional track plan.

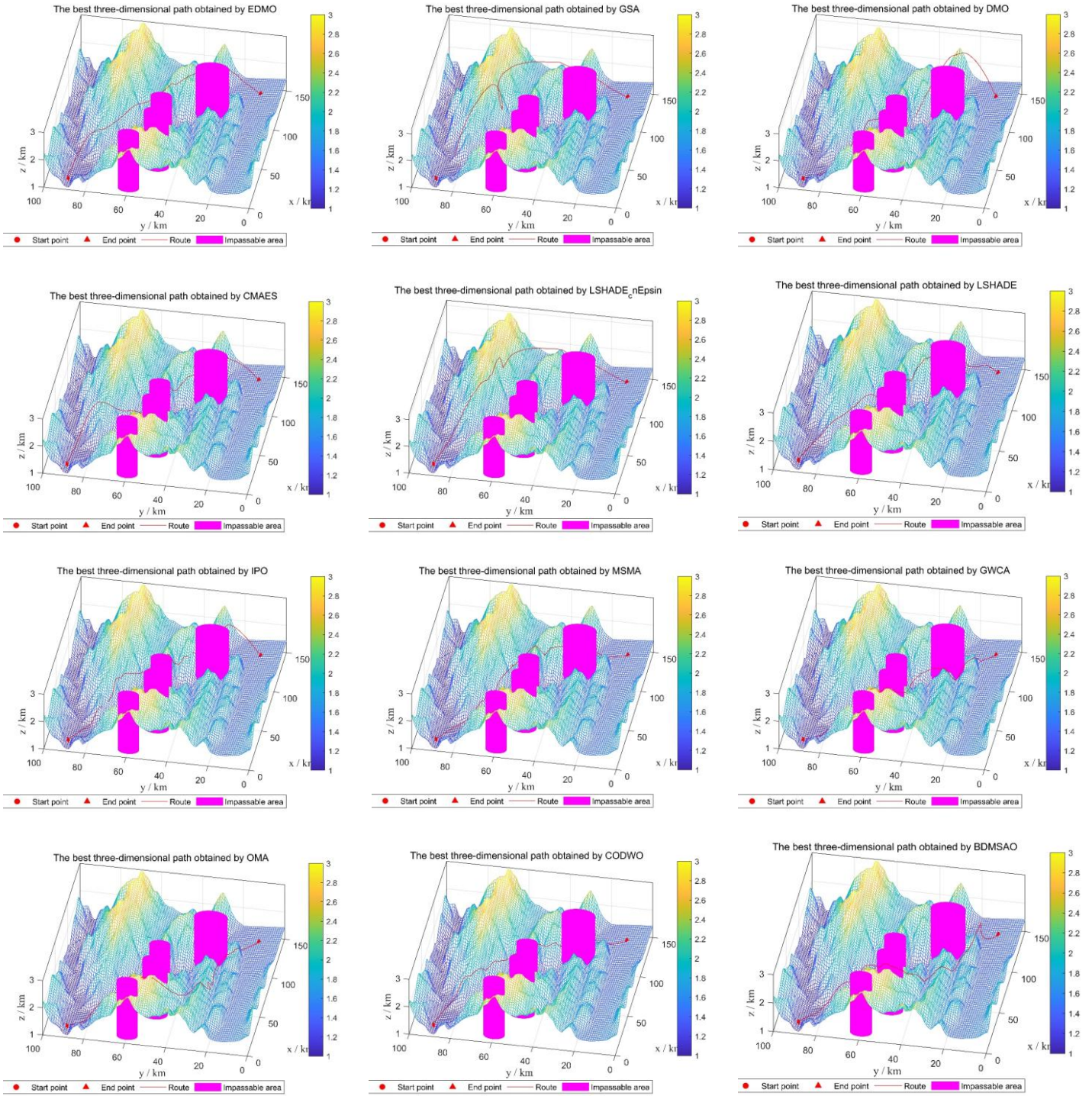

Figure 12. Three-dimensional track plan.

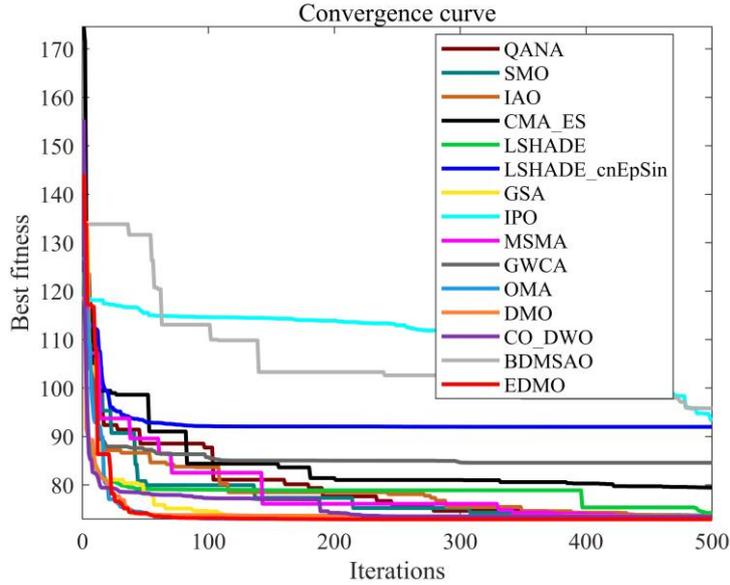

Figure 13. convergence curves.

### 6.3.2 Analysis of simulation result

The analysis of the experimental results reveals that the flight paths planned by the DMO algorithm are longer, leading to higher fuel consumption. More critically, these flight trajectories are in close proximity to risk areas, posing significant safety hazards and increasing the likelihood of flight accidents. In contrast, the optimized EDMO algorithm successfully mitigates these issues. It not only shortens the flight distance but also effectively avoids risk areas, significantly reducing the risks during mission execution. Furthermore, the improved EDMO algorithm plans for lower flight altitudes, allowing drones to navigate closer to mountainous terrain and effectively avoiding the flight risks associated with high-altitude cold temperatures and low oxygen levels.

Analysis of the initial best fitness values indicates that the EDMO algorithm demonstrates the lowest starting best fitness value and the fastest convergence rate. This suggests that the improved EDMO not only has a more advantageous starting point but also reaches the target solution more efficiently compared to other algorithms, highlighting its exceptional optimization capabilities. In particular, the incorporation of the DQTOS significantly enhances the algorithm's ability to escape local optima during the early search stages, thereby contributing to a faster and more robust convergence. Especially in the context of 3D path planning for drones, the enhanced EDMO algorithm, by integrating multiple strategies including quantum tunneling, exhibits superior performance, thus validating the effectiveness and necessity of the proposed enhancement mechanisms.

Under the same testing conditions, this study conducted 30 independent simulation experiments for 15 different algorithms. Additionally, a statistical analysis of the comprehensive cost models for these algorithms was performed, with the relevant statistical results presented in Table 12. Detailed information on the runtime statistics can be found in Table 13.

Table 12. Statistics of UAV three-dimensional path planning results.

| Algorithm | Best Cost | Worst Cost | Average Cost | Standard Deviation | Initial best fitness |
|---|---|---|---|---|---|
| **EDMO** | **72.91** | **72.96** | **72.95** | **0.03** | **152.55** |
| GSA | 79.97 | 86.07 | 85.02 | 0.18 | 214.62 |
| DMO | 73.01 | 73.21 | 73.12 | 0.09 | 240.58 |
| CMA-ES | 72.86 | 72.91 | 72.88 | 0.06 | 174.44 |
| LSHADE_cnEpsin | 76.97 | 88.18 | 82.11 | 0.58 | 171.75 |
| LSHADE | 72.86 | 73.10 | 72.91 | 0.07 | 172.68 |
| IPO | 74.02 | 79.06 | 76.75 | 0.15 | 224.34 |
| MSMA | 74.87 | 80.67 | 77.47 | 0.17 | 240.53 |
| GWCA | 74.55 | 78.04 | 76.86 | 0.09 | 174.46 |
| OMA | 78.12 | 83.41 | 81.56 | 0.18 | 185.63 |
| CO_DWO | 73.96 | 78.45 | 75.43 | 0.07 | 184.75 |
| BDMSAO | 79.05 | 85.87 | 83.44 | 0.45 | 214.43 |
| QANA | 75.68 | 79.32 | 77.59 | 0.12 | 176.46 |
| SMO | 74.59 | 78.55 | 75.60 | 0.09 | 180.25 |
| IAO | 76.66 | 80.88 | 78.65 | 0.16 | 174.59 |

Table 13. Run time statistical analysis.

| Algorithm | Min Run Time (s) | Max Run Time (s) | Average Run Time (s) | Standard Deviation | Wilcoxon+/=/- |
|---|---|---|---|---|---|
| EDMO | 35.35 | 41.23 | 37.39 | 1.54 | = |
| GSA | 36.34 | 43.34 | 39.78 | 2.13 | + |
| DMO | 38.35 | 43.46 | 42.65 | 1.66 | - |
| CMA-ES | 26.64 | 29.43 | 25.74 | 1.62 | + |
| LSHADE_cnEpsin | 21.68 | 28.24 | 26.40 | 2.78 | - |
| LSHADE | 22.65 | 29.46 | 23.44 | 2.18 | - |
| IPO | 38.53 | 47.58 | 43.54 | 2.99 | - |
| MSMA | 37.35 | 42.45 | 39.67 | 1.85 | + |
| GWCA | 32.57 | 38.74 | 35.34 | 1.69 | - |
| OMA | 38.97 | 46.45 | 42.78 | 2.51 | - |
| CO_DWO | 36.86 | 43.34 | 39.32 | 2.37 | + |
| BDMSAO | 46.64 | 72.64 | 68.87 | 1.86 | + |
| QANA | 39.58 | 62.61 | 58.44 | 2.21 | + |
| SMO | 36.64 | 42.56 | 38.89 | 1.94 | + |
| IAO | 34.55 | 41.55 | 36.51 | 1.68 | + |

After comparing the results of the 15 experimental groups, it was observed that while the optimal and worst costs of the 15 algorithms in drone path planning were relatively similar, the EDMO algorithm demonstrated superior average performance, ranking just slightly behind CMA-ES and LSHADE.

Despite their strong performance, CMA-ES and LSHADE have notable limitations. CMA-ES, while effective in unimodal and continuous optimization problems, struggles with high-dimensional multimodal landscapes due to its covariance matrix adaptation, which can slow down convergence and limit exploration diversity. Similarly, LSHADE, although adaptive in parameter control, often suffers from slower convergence rates, making it less efficient in dynamic UAV path planning where real-time adaptability is crucial. Other algorithms, such as GWO and SSA, exhibit high variability in solution quality, as their reliance on stochastic operators leads to inconsistent performance, particularly in constrained three-dimensional environments.

A key advantage of EDMO lies in its significantly lower initial best fitness value compared to other algorithms, suggesting that its search process starts closer to the global optimum, thereby reducing the likelihood of being trapped in local optima. This improvement is attributed not only to EDMO's OLOBL strategy, which enable more effective early-stage exploration and ensure rapid convergence toward high-quality solutions, but also to the integration of the DQTOS. By facilitating probabilistic transitions across potential barriers, this strategy greatly enhances the algorithm's capability to escape local optima and maintain global exploration efficiency. Furthermore, the BDFSS adaptively refines the search direction by simulating phototactic behavior, allowing the algorithm to dynamically focus on promising regions and significantly improving local exploitation precision. Unlike CMA-ES and LSHADE, EDMO maintains a superior balance between exploration and exploitation, preventing stagnation while consistently improving solution accuracy. In UAV applications, where predictability and reliability are crucial for efficient mission planning and risk management, the enhanced stability and adaptability endowed by EDMO make it a more suitable choice for three-dimensional drone path planning, particularly in complex and dynamic environments requiring robust and computationally efficient trajectory optimization.

Table 13 not only presents the minimum, maximum, standard deviation, and mean of the runtime but also includes a statistical analysis of the results using the Wilcoxon rank-sum test. This test is employed to assess the significance of the runtime between EDMO and other algorithms at a 5% significance level. A significance level of 5% indicates that there is a significant difference in runtime efficiency between the two algorithms; otherwise, the difference is not significant. The rank-sum test evaluates the test results, where "+", "-", and "=" denote that the significance level of EDMO is greater than, less than, or equal to 5%, respectively. As shown in Table 13, EDMO demonstrates a faster average runtime, ranking just below LSHADE_cnEpsin, LSHADE, GWCA and IAO. Notably, EDMO has the lowest standard deviation, indicating high consistency in the algorithm's performance across different runs. This suggests that during repeated tests, the runtime of the algorithm exhibits minimal fluctuations, reflecting greater stability and reliability.

The results of the Wilcoxon rank-sum test indicate that the EDMO algorithm exhibits statistically significant differences in runtime compared to DMO, LSHADE_cnEpsin, LSHADE, IPO, GWCA, and OMA. This means that, in terms of runtime performance, EDMO is significantly different from these algorithms. In contrast, the differences between EDMO and the other algorithms are not significant, suggesting that their efficiency in solving the 3D path planning problem for drones is comparable. Overall, EDMO not only performs well in terms of runtime but also demonstrates high stability and consistency across multiple runs. While it may be comparable to certain algorithms in terms of efficiency, it likely holds an advantage in the stability of runtime.

# 7 Conclusion and prospect of future work

This study systematically addresses the computational limitations and structural constraints of conventional DMO algorithm by proposing an enhanced framework that integrates three synergistic strategies. First, an OLOBL strategy is introduced to decouple high-dimensional feature correlations via Gram-Schmidt orthogonalization. Second, a DQTOS is employed, which leverages time-variant potential modeling to effectively overcome energy barriers associated with local optima. Third, a BDFSS is developed, drawing from microbial light-responsive navigation and utilizing an adaptive fitness-to-illuminance mapping function. Extensive experimental evaluations on the CEC2017 and CEC2020 benchmark suites demonstrate that the proposed framework achieves a robust balance between exploration and exploitation in non-convex landscapes and outperforms existing methods in complex multi-objective optimization tasks.

The performance of the EDMO algorithm is evaluated using the CEC2017 and CEC2020 test suites. The results indicate that the EDMO algorithm offers greater reliability in terms of convergence and outperforms other well-known algorithms in global optimization. The findings clearly demonstrate that EDMO achieves an optimal balance between exploration and exploitation, statistically surpassing other benchmark algorithms. In practical applications, particularly in engineering tasks and 3D path planning for drones, EDMO shows outstanding performance compared to algorithms such as QANA, SMO, IAO, CMA-ES, LSHADE_cnEpsin, LSHADE, DMO, GSA, IPO, MSMA, GWCA, OMA, CO_DWO, and BDMSAO, showcasing its exceptional optimization capabilities. This underscores the algorithm's significant potential in addressing complex engineering challenges.

Future research should focus on developing binary and multi-objective variants of the EDMO algorithm to conduct a comprehensive performance evaluation. This theory could be applied to problems such as optimizing renewable energy systems, wind farm layouts, photovoltaic models, and parameter estimation for fuel cells. Additionally, the algorithm may be utilized to address combinatorial optimization problems like logistics delivery route optimization and production scheduling, as well as vehicle path optimization, traffic signal control, and fleet scheduling in intelligent transportation systems. Through these applications, the potential of the EDMO algorithm across various fields can be further explored, enhancing its value in solving complex optimization problems.

## Compliance with Ethical Standards

### Author Contributions
Mingyang Yu: Conceptualization, Methodology, Writing - original draft, Formal analysis, Data curation, Writing - review & editing, Software. Haorui Yang: Visualization, Formal analysis, Writing - review & editing. Jing Xu: Conceptualization, Resources, Supervision, Formal analysis. Xiaoxuan Xu: Visualization, Resources, Writing - review & editing. Kangning An: Software, Writing - review & editing, Resources. Xinjian Wei: Methodology, Visualization, Resources, Software.

### Funding
This work was supported by Natural Science Foundation of Tianjin Municipality (21JCYBJC00110).

### Data Availability
All data generated or analysed during this study are included in this published article.

### Conflict of Interest
The authors declare that they have no conflict of interest.

### Competing Interests
The authors have no relevant financial or non-financial interests to disclose.

### Ethical Approval
His article does not contain any studies with human participants or animals performed by any of the authors.

### Informed Consent
This article does not contain any studies with human participants. So informed consent is not applicable here.

### Acknowledgement
There is no acknowledgement involved in this work.